%% file: main_arxiv.tex
\documentclass{article}
\usepackage[square,sort,comma,numbers]{natbib}
\usepackage[margin=1in,footskip=0.25in]{geometry} 

\usepackage[utf8]{inputenc} 
\usepackage[T1]{fontenc}    
\usepackage{hyperref}       
\usepackage{url}            
\usepackage{booktabs}       
\usepackage{amsfonts}       
\usepackage{nicefrac}       
\usepackage{microtype}      
\usepackage{amsfonts}
\usepackage{amssymb}
\usepackage{graphicx}
\usepackage{verbatim}
\usepackage{authblk}
\usepackage{graphicx}
\usepackage{doi}
\usepackage{times}
\usepackage{amsthm}
\usepackage{mathrsfs}
\usepackage{authblk}
\usepackage{amsmath,amsfonts,amssymb, graphicx, url} 

\hypersetup{colorlinks,citecolor=blue,linkcolor=blue,urlcolor=blue}  
\usepackage[nameinlink,capitalise]{cleveref}

\usepackage{url}            
\usepackage{booktabs}       
\usepackage{amsfonts}       
\usepackage{nicefrac}       
\usepackage{microtype}      
\usepackage[framemethod=TikZ]{mdframed}
\mdfsetup{skipabove=10pt,skipbelow=5pt,roundcorner=4pt}

\definecolor{light-gray}{gray}{0.95}
\usepackage[most]{tcolorbox}
\newtcbox{\mymath}[1][]{%
    nobeforeafter, math upper, tcbox raise base,
    enhanced, colframe=blue!30!black,
    colback=blue!30, boxrule=1pt,
    #1}

\usepackage{enumitem}
\usepackage{wrapfig}

\usepackage{verbatim}
\usepackage[titletoc]{appendix}
\usepackage{booktabs} 
\usepackage{multirow}
\usepackage{algorithm}
\usepackage[noend]{algpseudocode}
\usepackage{color}
\usepackage{amsthm}
\usepackage{esvect}
\usepackage{xcolor}
\usepackage{mathtools,amssymb,lipsum}

\usepackage{cuted}
\usepackage{relsize}

\algnewcommand{\LineComment}[1]{\State \(\triangleright\) #1}
\usepackage{tikz}
\usetikzlibrary{decorations.pathreplacing,calc}

\usepackage{subcaption}
\usepackage{graphicx}
\usepackage{siunitx}
\usepackage{caption}
\usepackage{booktabs}
\usepackage{makecell}

\usepackage[T1]{fontenc}
\usepackage[font=small,labelfont=bf,tableposition=top]{caption}

\DeclareCaptionLabelFormat{andtable}{#1~#2  \&  \tablename~\thetable}

\newtheorem*{rep@theorem}{\rep@title}
\newcommand{\newreptheorem}[2]{%
\newenvironment{rep#1}[1]{%
 \def\rep@title{#2 \ref{##1}}%
 \begin{rep@theorem}}%
 {\end{rep@theorem}}}

\newtheorem{proposition}{Proposition}

\newtheorem{definition}{Definition}

\newreptheorem{theorem}{Theorem}
\newreptheorem{claim}{Claim}

\usepackage{multirow}
\usepackage{array}
\usepackage{url}

\usepackage[sc,osf]{mathpazo}

\usepackage{xcolor}

\usepackage{adjustbox}

\newcommand{\textb}[1]{\textcolor{blue}{#1}}

\newif\ifdebug
\debugtrue

\ifdebug
    \newcommand{\citeb}[1] {\textcolor{blue}{\cite{#1}}}

\else
    \newcommand{\citeb}[1]{\cite{#1}}
\fi

\usepackage{soul}

\usepackage{hyperref}
\usepackage{xcolor}
\usepackage{multirow}

\usepackage{adjustbox}

\makeatletter
\newcommand*\bigcdot{\mathpalette\bigcdot@{.4}}
\newcommand*\bigcdot@[2]{\mathbin{\vcenter{\hbox{\scalebox{#2}{$\m@th#1\bullet$}}}}}
\makeatother

\newtheorem*{proposition*}{Proposition}

\title {Bayesian Convolutional Deep Sets with Task-Dependent Stationary Prior}
\author{Yohan Jung, Jinkyoo Park \\ KAIST }

\date{}

\begin{document}

\maketitle

\input{00-abstract-v02}

\input{01-introduction-v02}

\input{02-preliminaries-v01}

\input{03-methodology-v06}

\input{05-experiments-v05}
\input{04-relatedwork-v05} 

\clearpage
\bibliographystyle{unsrt}
\bibliography{citation}


\end{document}

%% file: 00-abstract-v02.tex
\begin{abstract}

Convolutional deep sets are the architecture of a deep neural network (DNN) that can model stationary stochastic process. This architecture uses the kernel smoother and the DNN to construct the translation equivariant functional representations, and thus reflects the inductive bias of the stationarity into DNN. However, since this architecture employs the kernel smoother known as the non-parametric model, it may produce ambiguous representations when the number of data points is not given sufficiently. To remedy this issue, we introduce Bayesian convolutional deep sets that construct the random translation equivariant functional representations with stationary prior. Furthermore, we present how to impose the task-dependent prior for each dataset because a wrongly imposed prior forms an even worse representation than that of the kernel smoother. We validate the proposed architecture and its training on various experiments with time-series and image datasets.
\end{abstract}

%% file: 01-introduction-v02.tex






\section{Introduction}

Neural process (NP) and Conditional neural process (CNP) \citeb{garnelo2018conditional,garnelo2018neural} are the pioneering deep learning framework for modeling stochastic process, i.e., the functions over the distribution. That is, for any finite input and output pairs referred as context sets, these NP models output the predictive distribution on targeted inputs (target sets) by extracting the feature from context sets. Specially, the NP models employ the Deep sets \citeb{zaheer2017deep} to reflect the exchangeability of the stochastic process into a predictive distribution of the NP. Many variants of NP \citeb{singh2019sequential,yoon2020robustifying,louizos2019functional,gordon2019convolutional,foong2020meta,bruinsma2021gaussian,kawano2020group,holderrieth2021equivariant} have been proposed to model stochastic process elaborately.

Some NP models impose a certain inductive bias to model the stochastic process having the structured characteristics. For example Convolutional conditional neural process (ConvCNP) \citeb{gordon2019convolutional} is a NP model designed for modeling stationary process of which statistical characteristics over the finite subset of the process, such as the mean and covariance, do not change even when the time indexes of those finite random variables are shifted. ConvCNP employs the Convolutional Deep sets (ConvDeepsets) that constructs the functional representation for stationary process, and thus reflects the inductive bias of stationarity on ConvCNP.

To construct the translation equivariant representation, ConvDeepsets employs the RBF kernel function and convolutional neural network (CNN). Specifically, ConvDeepsets first constructs the discretized functional representation of the context set by using the Nadaraya–Watson kernel smoother \citeb{nadaraya1964estimating}, and then maps the discretized representation to the abstract representation via CNN. Since the kernel smoother produces a consistent representation regardless of the translation of the inputs for the context sets, and the convolution operation preserves the translation equivariance, the corresponding representation could be used to make the predictive distribution for modeling the stationary process. However, since the kernel smoother is a non-parametric model whose expressive power is dependent on the amount of the given context set, the representation of the kernel smoother could be ambiguous when the number of context data points is not given sufficiently. This can result in the poor performance of the corresponding NP models because the ConvDeepsets can not produce a proper representation for modeling the target set. This is analogous to the task ambiguity issue \citeb{finn2018probabilistic} noted in model agnostic meta learning (MAML) \citeb{finn2017model}.

One intuitive approach to attenuate the task ambiguity is to introduce a reasonable prior distribution on the representation for the kernel smoother. In fact, the Bayesian approach, which imposes prior distribution on the model parameters, has shown meaningful results for tackling the task ambiguity in MAML \citeb{finn2018probabilistic}. However, using the prior distribution also raises the question of which prior distribution should be used. Extremely, if a wrong prior distribution is assumed for the given datasets, the assigned prior distribution may affect the representations and the outputs of NP model negatively.

In this work, we propose the Bayesian convolutional Deep sets that constructs random functional representations via a task-dependent stationary prior. To this end, we first consider a set of stationary kernel, each of which is characterized by its distinct spectral density. Then, we construct the task-dependent prior by using an amortized latent categorical variable that is modeled by the translate-invariant neural network; the latent variable assigns a proper kernel out of the candidate set depending on the task. Next, we construct the sample functions of the Gaussian process (GP) posterior using the chosen kernel, and forwarding those sample functions by CNN, which is a representation of a Bayesian ConvDeepsets. We ensure that Bayesian ConvDeepsets still satisfies the translation equivalence that is necessary for modeling the stationary process. 

For training, we employ the variational inference, and consider additional regularizer that allows the neural network to chose the stationary prior reasonably depending on the task. We validate that the proposed method relaxes the task ambiguity issue by assigning a task-dependent prior on the time-series and the image datasets. Our contributions can be summarised as:
\begin{itemize}[leftmargin=1em]
    \vspace{-1.5mm}
    \item  We propose the Bayesian ConvDeepsets using a task-dependent stationary prior and its inference to attenuate the potential task ambiguity issue of the ConvDeepsets.
    \vspace{-1.5mm}
    \item We validate that the Bayesian ConvDeepsets can improve the modeling performance of the NP models on various tasks of the stationary process modeling such as prediction of the time-series and spatial dataset.
\end{itemize}
\color{black}

\vspace{-2mm}

%% file: 02-preliminaries-v01.tex





%


\section{Preliminaries}
\label{sec:preliminary}
\paragraph{Neural Process.} NP uses Deepsets to reflect the exchangeability of the stochastic process into the predictive distribution of the NP, and employs the meta-learning for training.

Let  $X^{c}{=}\{x^{c}_{n}\}_{n=1}^{N^{c}}$ and $Y^{c}{=}\{y^{c}_{n}\}_{n=1}^{N^{c}}$ be the $N^{c}$ pairs of context inputs and outputs, and $D^c=\{X^{c},Y^{c}\}$ be the context set. Similarly, let  $X^{t}{=}\{x^{t}_{n}\}_{n=1}^{N^{t}}$, $Y^{t}{=}\{y^{t}_{n}\}_{n=1}^{N^{t}}$, and $D^t=\{X^{t},Y^{t}\}$ be the $N^{t}$ pairs of the target inputs, outputs, and target set. Then, NP trains the mapping parameterized by neural network $f_{\Theta_{\mathrm{nn}}}$ that maps the context set $D^c$ and the target inputs $X^{t}$ to the parameters of the predictive distribution, $\mu(X^{t})$ and $\sigma(X^{t})$, on target input $X^{t}$, i.e.,
\begin{align}
f_{\Theta_{\mathrm{nn}}}:\left( D^c, X^{t} \right) \longmapsto \left( \mu_{\text{nn}}(X^{t})  ,  \sigma_{\text{nn}}(X^{t}) \right)
\label{eqn:nppred}
\end{align}
by optimizing the following objective :
 \begin{align}
\max_{\Theta_{\text{nn}}} \ \mathbb{E}_{D^{c},D^{t} \sim p(\mathcal{T})} \bigl[ \log{p \left( Y^{t}| f_{\Theta_{\text{nn}} } \left( X^{t},D^c \right) \right) }  \bigr]
\end{align}
where $p(\mathcal{T})$ denotes the distribution of the task for the context set $D^{c}$ and target set $D^{t}$.


\paragraph{Translation Equivariance.} Stationary process is characterized by the property that its statistical characteristics do not change even when the time is shifted. Thus, functions that can model the stationary process satisfy the special conditions, referred as Translation Equivariance (TE). Mathematically, TE can be defined as follows:
\begin{definition} [\citeb{gordon2019convolutional}]
\label{def:TE}
Let $\mathcal{X}= R^{d}$ and $\mathcal{Y} \subset R^{d'}$ be space of the inputs and outputs, and let $\mathcal{D}=\cup_{m=1}^{\infty} (\mathcal{X}\times \mathcal{Y})^{m}$ be the joint space of the finite observations. Also, let $\mathcal{H}$ be the function spaces on $\mathcal{X}$, and $T$ and $T^{*}$ be the mappings:
\begin{align}
& T:\mathcal{X} \times \mathcal{\mathcal{D}},  \hspace{9.4mm} T_{\tau}(D) = ((x_1+\tau,y_1),..,(x_n+\tau,y_n)) \nonumber \\
& T^{*}:\mathcal{X} \times \mathcal{H},  \hspace{6mm} T^{*}_{\tau}(h(\bigcdot)) = h(\bigcdot-\tau) \nonumber  \nonumber
\end{align}
where $D=\{(x_n,y_{n})\}_{n=1}^{N} \in \mathcal{D}$ denotes $N$ pairs of the inputs and outputs, $\tau \in \mathcal{X}$ denotes translation variable for the inputs, and $h(\bigcdot) \in \mathcal{H}$ denote the function for any input $\bigcdot \in \mathcal{X}$. Then, a functional mapping $\Phi:\mathcal{D} \rightarrow \mathcal{H}$ is translation equivariant if the following holds:
\begin{align}
\Phi \circ (T_{\tau}(D)) =  T^{*}_{\tau} \circ (\Phi(D)).
\label{eqn:def1}
\end{align}
\end{definition}

Roughly, speaking \cref{def:TE} implies that the function satisfying the TE should produce a consistent functional representation up to the order of translation.

\vspace{-1mm}


\paragraph{Convolutional Deep Sets.} ConvDeepsets is the specific architecture of the neural network satisfying the TE in \cref{eqn:def1}, and thus can be used to model stationary process. The following proposition introduces the specific structure of the ConvDeepsets $\Phi(D) (\bigcdot)$.

\begin{proposition} [\citeb{gordon2019convolutional}]
\label{pro:convdeepset}
Given dataset $D=\{(x_n,y_n)\}_{n=1}^{N}$, its functional representation $\Phi(D)(\bigcdot)$ is translation equivariant if and only if $\Phi(D)(\bigcdot)$ is represented as
\vspace{-1mm}
\begin{align}
    \hspace{-0.5mm} 
       & \underbrace{E(D) (\bigcdot) }_{ \substack{ \mathrm{functional} \\ \mathrm{representation} }    } =
    \Big[ \  
           \underbrace{\sum_{n=1}^{N}  k(\bigcdot -  x_n)}_{\mathrm{density}} \hspace{1mm}, \hspace{1mm}
           \underbrace{\sum_{n=1}^{N}  
           \frac{y_n \ k(\bigcdot - x_n)}{ \sum_{n=1}^{N}  k(\bigcdot -  x_n)}      }_{\mathrm{data \ representation}}
          \
    \Big], \nonumber \\  
    & \underbrace{ \Phi(D) (\bigcdot) }_{ \substack{ \mathrm{ConvDeepsets} \\ \mathrm{representation} }    } =
    \underbrace{ \rho}_{ \substack{ \mathrm{mapping \ via} \\ \mathrm{ CNN} } } 
    \circ 
    \underbrace{E(D) (\bigcdot) }_{ \substack{ \mathrm{functional} \\ \mathrm{representation} }    }       
    \label{eqn:convdeepset}
\end{align}
where $k(\bigcdot \ - \ x_n)$ denotes the stationary kernel centered at $x_n$, and $\rho(\bigcdot)$ is the continuous and translation equivariant mapping. Here, the RBF kernel function is used, and $\rho(\bigcdot)$ can be parameterized by CNN. 
\end{proposition}
\vspace{-2mm}


\paragraph{Neural Process with Convolutional Deepsets.}  ConvCNP \citeb{gordon2019convolutional} and ConvLNP \citeb{foong2020meta} are well-known NP models that can model stationary process by using ConvDeepsets as the main structure of the NP model.

To employ the functional representation of ConvDeepsets in practice, these NP models first consider $M$ discretized inputs $\{t_m\}_{m=1}^{M} \subset [\min{X},\max{X}]$ by spacing the range of inputs $X=X^{c} \cup X^{t}$ linearly. Then, these models construct $M$ discretized functional representations $\{\Phi(D^{c})(t_m)\}_{m=1}^{M}$ on the discretized inputs $\{t_m\}_{m=1}^{M}$ with \cref{eqn:convdeepset} as,
\begin{align}
\Phi(D^{c}) (t_m) = \left( \rho \circ E(D^{c}) \right) (t_m) \hspace{4mm} m=1,..,M \label{eqn:finite_convdeepset} .
\end{align}
These discretized representations $\{\Phi(D^{c})(t_m)\}_{m=1}^{M}$ are used to obtain the parameters of the predictive distribution $\mu(X^{t})$ and $\sigma(X^{t})$ as shown in \cref{eqn:nppred}. Specially, the smoothed representations on target inputs $ x^{t}_{n} \in X^{t}$, i.e,
\begin{align}
\vspace{-1mm}
\sum_{m=1}^{M}  \Phi(D^{c})(t_m) \ k( x^{t}_{n} - t_m ) 
\label{eqn:smoothed}
\end{align}
are used for modeling predictive distribution $p(Y^{t} | X^{t},D^{c} )$. For the grid dataset, we can omit the discretization procedure and employ the CNN directly \citeb{gordon2019convolutional}.

%% file: 03-methodology-v06.tex

\section{Methodology}
\begin{figure*}[t!]
\centering
{\includegraphics[width=0.98\linewidth]{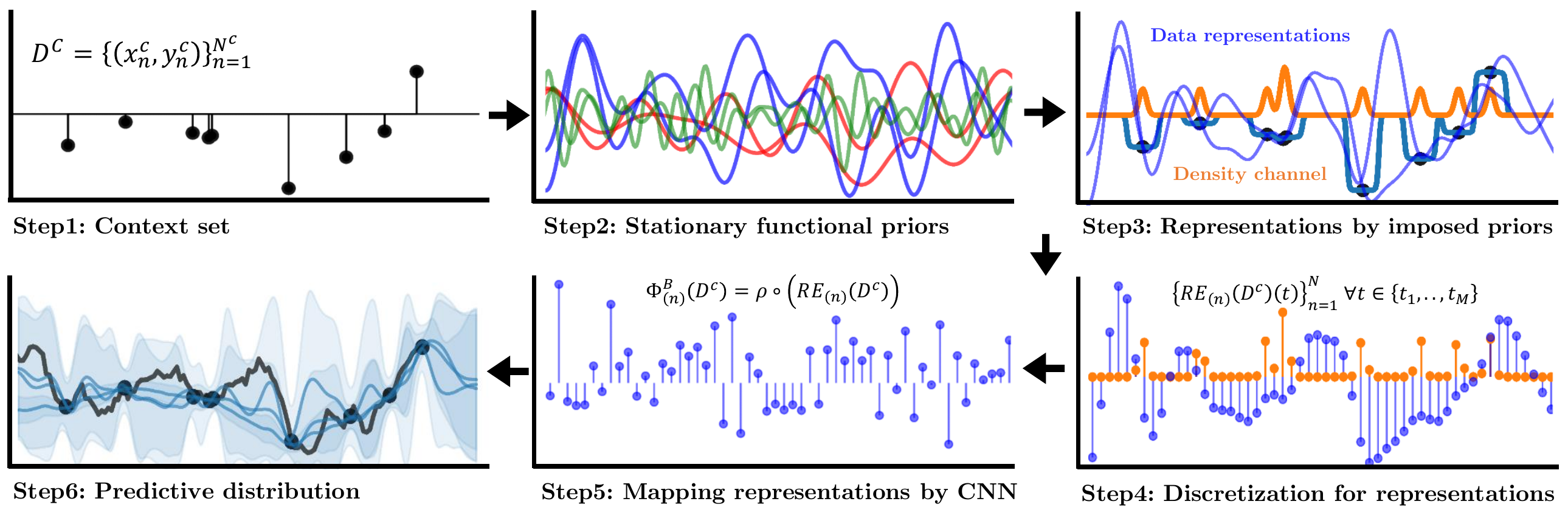}} 
\caption{Prediction procedure: \textbf{Step 1:} Given context set $D^{c}$. \textbf{Step 2:} Consider the random stationary functions $\{\phi_{q}\}_{q=1}^{Q}$ of \cref{eqn:mixed_fourier}. \textbf{Step 3:} Assign the task-dependent kernel by \cref{eqn:latent_sample,eqn:mixedkernel} out of stationary priors in Step 2, and construct $N$ random representations $\{RE_{(n)}(D^{c}) (\bigcdot)\}_{n=1}^{N}$ in \cref{eqn:bayes_convdeepset}. \textbf{Step 4:} Discretize the representation as explained in \cref{eqn:finite_convdeepset}. \textbf{Step 5:} Map the discredited representation via CNN, i.e., ${\Phi^{\mathcal{B}}_{(n)}}(D^{c})(\bigcdot)$. \textbf{Step 6:} Smooth the mapped representations by \cref{eqn:smoothed}, and make prediction distribution by \cref{eqn:prediction}. For the Step 4 and 5, one random representation is depicted for clarity.}
\label{fig:main-illustration2}
\end{figure*}

In this section, we first interpret the representation of the ConvDeepsets and its motivation in \cref{subsec:method1}. Then, we introduce the task-dependent stationary prior in \cref{subsec:method2}, the Bayesian ConvDeepsets in \cref{subsec:method3}, and its application to stationary process modeling in \cref{subsec:method4}.
\cref{fig:main-illustration2} outlines the prediction procedure via Bayesian ConvDeepsets described in \cref{subsec:method2,subsec:method3,subsec:method4}.

\clearpage

\subsection{Interpretation of ConvDeepsets Representation} 
\vspace{2mm}
\label{subsec:method1}

\begin{wrapfigure}[27]{r}{7.5cm}
\centering
\vspace{-1mm}
\subfloat[\label{fig:concept-1} Data representation of ConvDeepsets]
{\includegraphics[width=0.95\linewidth]{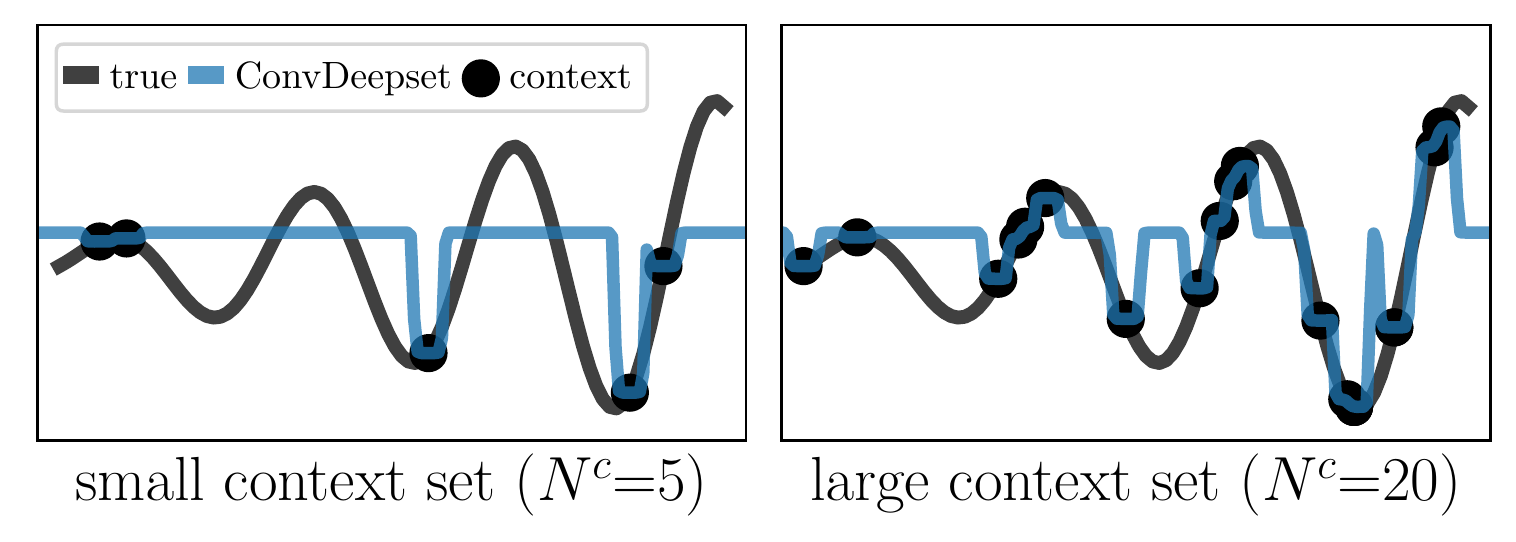}} \vspace{1mm}

\subfloat[\label{fig:concept-2} Data representation with stationary functional prior]
{\includegraphics[width=0.95\linewidth]{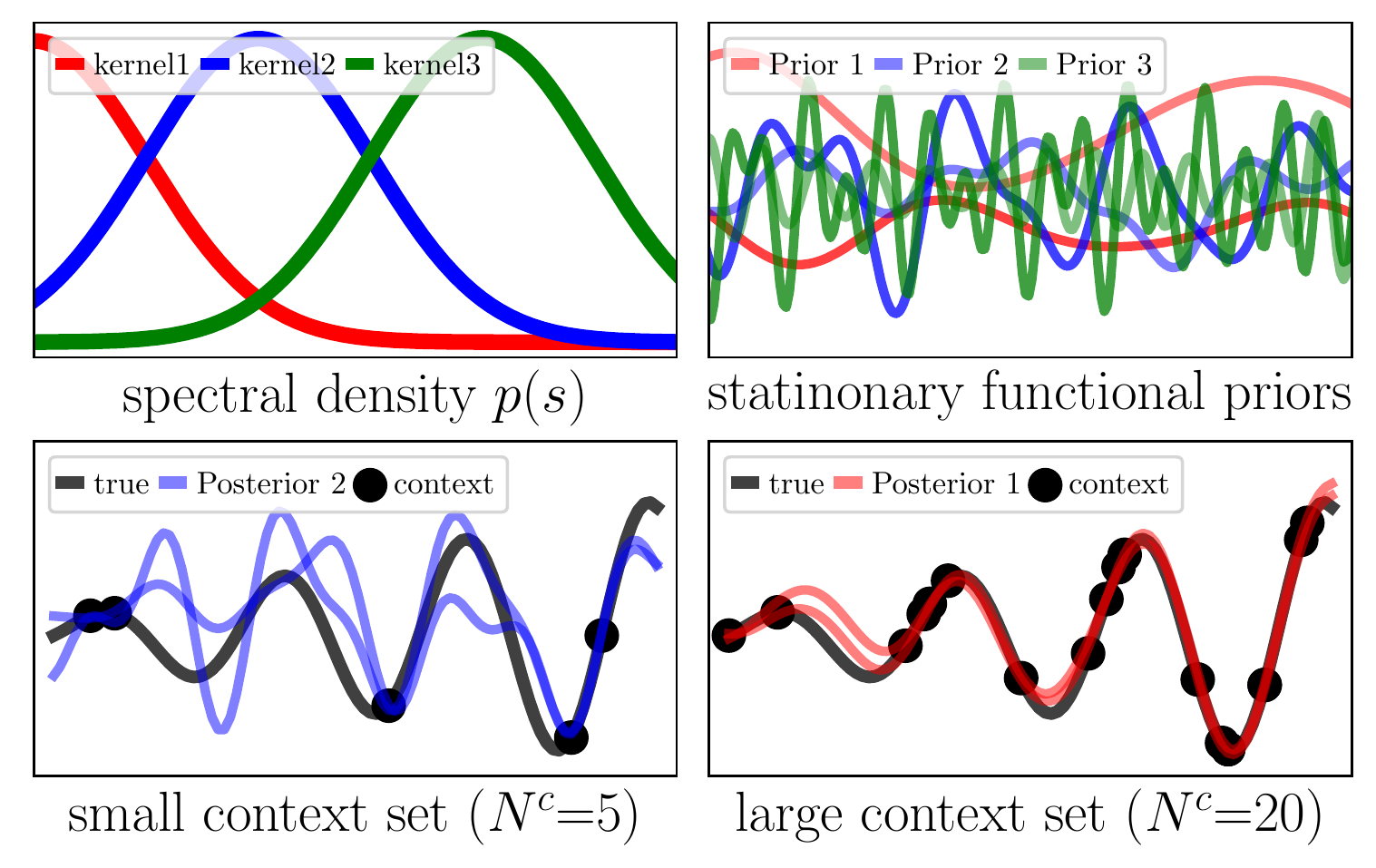}}   
\caption{Comparison of data representations over different $N^{c}$ context points. \cref{fig:concept-1} shows the data representation of $E(D^{c})$ for ConvDeepsets; the black dots denotes the context set $D^{c}$, and black and blue line denote the true function and the data representation, respectively. \cref{fig:concept-2} shows the stationary priors characterized by each spectral density $p_{q}(s)$ in \cref{eqn:basiskernel}, and the improved data representation due to the reasonably imposed prior, especially for the $N^{c}=5$.
}
\label{fig:concept}
\end{wrapfigure}
\vspace{-2mm}

The data representation of $E(D^{c}) (\bigcdot)$ in \cref{eqn:convdeepset} is constructed by the RBF kernel smoother known as the non-parametric model. The lengthscale of the RBF kernel controls the smoothness of the data representation; when the lengthscale is close to 0, the data representation is expressed as $\sum_{n=1}^{N_c} y^{c}_n  \ \delta (\bigcdot -  x^{c}_n )$ with a Dirac-delta function $\delta(\bigcdot)$. This implies that the expressive power of the kernel smoother is proportional to the number of context points $N^c$. When small $N^{c}$ context data is given, the data representation could be the ambiguous, and thus affect $\Phi(D^{c})$ and the predictive distribution $p(Y^{t} | X^{t},D^{c} )$ negatively.


To remedy the issue, we note that the data representation of $E(D^{c})(\bigcdot)$ in \cref{eqn:convdeepset} can also be expressed as the re-scaled predictive mean of GP posterior with the restricted covariance
$\text{Diag}(K(X^{c},X^{c}))$, as follows:
\vspace{-2mm}
\begin{align} 
&\sum_{n=1}^{N} \frac{y_n \ k(\bigcdot - x_n)}{ \sum_{n=1}^{N}  k(\bigcdot -  x_n) }  \nonumber \\
&= \frac{1}{\sum_{n=1}^{N^{c}}  k(\bigcdot -  x^{c}_n)} 
\underbrace{ K(\bigcdot,X^{c}) \text{Diag}(K(X^{c},X^{c}))^{-1} Y^{c} }_{\mathrm{predictive \ mean \ of \ GP \ posterior}}  \label{eqn:interpretation}
\end{align}
where $[\text{Diag}(K(X^{c},X^{c}))]_{n,m} =  1_{n=m}$ and $K(\bigcdot,X^{c}) \in R^{1 \times N^{c}}$. Based on this observation, we hypothesis that (1) setting a reasonable stationary GP prior depending on each task, and (2) using or adding the sample functions of the GP posterior to represent the data part in $E(D^{c})(\bigcdot)$ could alleviate the unclear representation of the ConvDeepsets, when using small $N^{c}$ context data points. \cref{fig:concept} describes our concern of the data representation for $E(D^c)$, and its intuitive approach to attenuate this issue.

\vspace{-2mm}
\subsection{Amortized Task-Dependent Stationary Prior}
\label{subsec:method2}

To impose the reasonable prior depending on the task, we employ an amortized latent variable approach for imposing the stationary prior. To this end, based on the Bochner's theorem \citeb{} that explains how the stationary kernel can be constructed by specifying the spectral density, we construct $Q$ stationary kernels $\{k_q\}_{q=1}^{Q}$ as follows:
\begin{align}
k_q(\tau) = \int e^{i 2\pi   s^{T} \tau}p_q(s) ds
\label{eqn:basiskernel}
\end{align}
where $p_q(s)$ denotes $q$-th spectral density defined on frequency domain. In this work, we model the spectral density  $p_q(s)=N(s;\mu_q,\mathrm{Diag}(\Sigma_q))$, and order the $\{k_q\}_{q=1}^{Q}$ by $\mu_1{=}0$ and $\mu_1 \leq .. \leq \mu_Q$ in element-wise sense.

Then, we consider the categorical latent variable $Z_{\text{cat}} = (z_1,..,z_Q)$ parameterized by the neural network as follows:
\begin{align}
q(Z_{\text{cat}} \ | \ D^c) = \mathrm{Cat} \left( Z_{\text{cat}} \ ; \ \mathrm{p}_{\mathrm{nn}}(D^{c}) \right) \label{eqn:latentcat},
\end{align}
where $\mathrm{p}_{\mathrm{nn}}: D^c \longrightarrow {\Delta}^{Q-1} $, with $Q$-dimension simplex ${\Delta}^{Q-1}$, denotes the parameter of categorical variable parameterized by the translate invariant neural network satisfying 
\begin{align}
\vspace{-2mm}
\mathrm{p}_{\mathrm{nn}} (T_{\tau}(D^{c})) = \mathrm{p}_{\mathrm{nn}}(D^{c})  \label{eqn:transinv_net},
\end{align}
where $T_{\tau}(D^{c}) {=} \left((x^{c}_{1}+\tau,y^{c}_1), .., (x^{c}_{N^{c}}+\tau,y^{c}_{N^{c}}) \right) $ denotes the $\tau$-translated context set defined in \cref{def:TE}. This $\mathrm{p}_{\mathrm{nn}} $ is designed to make the derived functional representation satisfy the TE, which will be stated in \cref{pro:Bayes_convdeepset}. The structure of $\mathrm{p}_{\mathrm{nn}}$ in this work is described in Appendix A.2. 

Using the kernels $\{k_q(\tau) \}_{q=1}^{Q}$ in \cref{eqn:basiskernel} and the latent variable $Z_{\mathrm{cat}}(D^{c})$ in \cref{eqn:latentcat}, we build the amortized latent stationary kernel depending on the context set $D^{c}$ as
\begin{align}
k(\tau) \ | \ Z_{\text{cat}}(D^{c}) = z_1 k_1(\tau)  \hspace{2mm} +  \hspace{2mm} \cdots  \hspace{2mm}  + \hspace{2mm} z_Q k_Q(\tau).
\label{eqn:mixedkernel}
\end{align}
\paragraph{Multi-channel Extension.}  The latent stationary kernel in \cref{eqn:mixedkernel} can be extended for multi-channel processes ($K$ channels) by considering $K$-th channel categorical variables $\{Z_{\text{cat}}^{k}(D^{c})\}_{k=1}^{K}$ with $Z_{\text{cat}}^{k}(D^{c}) {\sim} \mathrm{Cat}( \mathrm{p}^k (D^{c}) )$, where $\mathrm{p}^k (D^{c})$ denotes the amortized parameter for $k$-th channel.
\vspace{-2mm}

\subsection{Bayesian Convolutional Deep sets.} 
\label{subsec:method3}
\vspace{-1mm}

Based on the latent stationary kernel in \cref{eqn:mixedkernel}, we build the random data representation, and the construct the Bayesian ConvDeepsets. To this end, we employ the path-wise sampling for GP posterior, that constructs the sample function of GP posterior efficiently; this can reduce the computation cost of conventional posterior sampling by computing the prior function and update function of the data separately \citeb{wilson2020efficiently}. Let $f(D^c)(\bigcdot)$ be the random data representation for the context set $D^c = \{(x^{c}_n,y^{c}_{n})\}_{n=1}^{N^c}$, and it can be represented
\vspace{-2mm}
\begin{align}
\hspace{-3mm}
\underbrace{ f(D^c)(\bigcdot) }_{ \substack{\mathrm{random \ data} \\ \mathrm{representation}} } & \coloneqq \hspace{3mm} 
\underbrace{\sum_{q=1}^{Q} \textb{ \sqrt{z_q} } \ \textb{\phi_q(\bigcdot)} }_{\mathrm{prior \  term}} 
\hspace{.5mm} + \hspace{.5mm} 
\underbrace{\sum_{n=1}^{N^{c}} \textb{v_{n}}  \ \textb{k(\bigcdot - x^{c}_{n}} )}_{\mathrm{update \ term }}.
\label{eqn:randomfunctionalfeature}
\end{align}

For computing the prior, we construct the random samples $(z_{1},..,z_{Q})$ of the latent categorical variable of \cref{eqn:latentcat}, i.e.,  
\begin{align}
\textb{(z_{1},..,z_{Q})}  \sim \mathrm{Cat} \left( Z_{\text{cat}} \ ; \ \mathrm{p}_{\mathrm{nn}}(D^{c}) \right)
\label{eqn:latent_sample}
\end{align}
by applying the Gumbel-softmax trick \citeb{jang2016categorical} to the output of $\mathrm{p}_{\mathrm{nn}}(D^{c})$. We construct the $q$-the random stationary function $ \phi_q(\bigcdot)$, i.e., a sample function of GP prior using $q$-th kernel $k_q$, based on random Fourier feature (RFF) \citeb{rahimi2008random,jung2022efficient}, as 
\vspace{-2mm}
\begin{align}
 & \{w_{q,i}\}_{i=1}^{l} {\sim} N(0,I), \ \{s_{q,i}\}_{i=1}^{l} {\sim} p_{q}(s), \  \{b_{q,i}\}_{i=1}^{l} {\sim} U[0,2\pi],   \nonumber \\
 & \hspace{5mm} \textb{\phi_q(\bigcdot)} = \sqrt{\frac{2}{l}} \sum_{i=1}^{l} w_{q,i} \cos{ ( 2 \pi \langle  s_{q,i}, \bigcdot \rangle + b_{q,i})} \label{eqn:mixed_fourier},   
\end{align}
where $w_{q,i}$ denotes the random weight samples, $s_{q,i}$ and $b_{q,i}$ denote the spectral points sampled from $p_{q}(s)$ and phase sampled from uniform distribution $U[0,2\pi]$, respectively.

For comping the update term for \cref{eqn:randomfunctionalfeature}, we compute the expected kernel for the latent kernel of \cref{eqn:mixedkernel} as
\vspace{-.5mm}
\begin{align}
\textb{ k(\bigcdot - x^{c}_{i} ) }
&= \mathbb{E}_{ q(Z_{\text{cat}} \ | \ D^{c} ) }
 [  k(\bigcdot - x^{c}_{i} ) \ | \ Z_{\text{cat}}(D^{c})  ]  \label{eqn:updateterm} 
\end{align}
where $Z_{\text{cat}}(D^{c})$ is integrated over $q(Z_{\text{cat}}  |D^{C})$. We compute the smoothing weight $v_{n} \in R$ for the $n$-th context input $x^{c}_{n}$,
\begin{align}
 \textb{v_{n}} &= \left[  K(X^{c},X^{c}) {+} \sigma^{2}_{\epsilon}I  ) {}^{-1}
          \left(Y^{c} - \Psi(X^{c}) \right)  \right]_{n},  \label{eqn:updateterm-2}  
\end{align}
where $ K(X^{c},X^{c}) \in R^{N^c \times N^c} $ denotes Gram matrix of the expected kernel $ k$ in \cref{eqn:updateterm},  and $Y^c \in {R^{N^c}}$ denotes the stacked vector of the context outputs $\{y^{c}_{n}\}_{n=1}^{N^c}$. $\Psi(X^{c}) \in {R^{N^c}} $ denotes the values of the random stationary function on $X^{c}$, which is computed by the prior term in \cref{eqn:randomfunctionalfeature}.

\cref{eqn:randomfunctionalfeature} can be also approximated, i.e., $f(D^c)(\bigcdot) \approx$
\begin{align}
\hspace{-.5mm}
\alpha
\underbrace{  \sum_{q=1}^{Q} \sqrt{z_q}   \phi_   q(\bigcdot)}_{\text{prior term}} +
\textb{\Tilde{k}} * 
\Big(
\underbrace{
\sum_{n=1}^{N^{c}} y^{c}_n   \delta (\bigcdot -  x^{c}_n ) 
}_{\text{data term}}
- 
\alpha
\underbrace{
\sum_{q=1}^{Q} \sqrt{z_q}  \phi_q(\bigcdot) 
}_{\text{prior term}}
\Big) \nonumber
\end{align}
for scalability, where $\alpha \in (0,1)$ denotes the hyperparameter for controlling how much the stationary prior is reflected on $f(D^c)(\bigcdot)$. $ \textb{\Tilde{k}}$ denotes the filter of convolution, and is set to truncate the expected kernel of \cref{eqn:updateterm} with the finite length (filter window size). This approximation scheme is designed to reflect the stationary prior and the data term on the data representation as $f(D^c)(\bigcdot)$ constructs the representation in \cref{eqn:randomfunctionalfeature}. We validate that this scheme alleviates the task ambiguity on the image completion task (\cref{subsec:exp3}).

Then, we can construct the random functional representation $RE(D) (\bigcdot)$ through the $f(D^{c})(\bigcdot)$ in \cref{eqn:randomfunctionalfeature}, and the Bayesian ConvDeepsets $\Phi^{\mathcal{B}} (D^{c}) (\bigcdot) $ as the ConvDeepsets $\Phi (D^{c}) (\bigcdot) $ is constructed in \cref{pro:convdeepset} as follows:  
\vspace{-1mm}
\begin{align}
    \hspace{-0.5mm} 
       \underbrace{RE(D^{c}) (\bigcdot) }_{ \substack{ \mathrm{random \ functional} \\ \mathrm{representation} }    }  & =
    \Big[ \  
      \underbrace{\sum_{n=1}^{N^{c}}  k(\bigcdot -  x^{c}_n)}_{\mathrm{density}} 
          \hspace{3mm}  , \ 
         \underbrace{f(D^c)(\bigcdot) }_{ \substack{ \mathrm{random \ data} \\ \mathrm{representation}} }
          \
    \Big], \nonumber \\ 
 \underbrace{  \Phi^{\mathcal{B}} (D^{c}) (\bigcdot) }_{ \substack{ \mathrm{Bayes \ ConvDeepsets} \\ \mathrm{representation} }    }   & =
    \underbrace{ \rho}_{ \substack{ \mathrm{mapping \ via} \\ \mathrm{ CNN} } } 
    \circ 
    \underbrace{RE(D^{c}) (\bigcdot) }_{ \substack{ \mathrm{random  \ functional} \\ \mathrm{representation} }    }       
    \label{eqn:bayes_convdeepset} 
\end{align}

We prove that the Bayesian ConvDeepsets still holds the translation equivariance (TE) in the following proposition.
\begin{proposition} 
\label{pro:Bayes_convdeepset}
Given dataset $D=\{(x_n,y_n)\}_{n=1}^{N}$, if the Bayesian ConvDeepsets $\Phi^{\mathcal{B}} (D) (\bigcdot)$ is defined on the finite grid points, $\Phi^{\mathcal{B}} (D) (\bigcdot)$ is still translation equivariant in distribution sense, i.e.,
\begin{align}
\Phi^{\mathcal{B}}  \circ (T_{\tau}(D)) \stackrel{d}{=}  T^{*}_{\tau} \circ (\Phi^{\mathcal{B}}(D)).
\label{eqn:te_probsense}
\vspace{-3mm}
\end{align}

\begin{proof}
The proof can be checked in Appendix A.3.
\end{proof}
\vspace{-2mm}
\end{proposition}

\vspace{-3mm}
\subsection{Prediction and Training}
\vspace{-1mm}
\label{subsec:method4}

\paragraph{Prediction.} For the context set $D^{c}$ and target set $D^{t}$, 
the Bayesian ConvDeepsets can be used for computing predictive distribution $p(Y^{t}| X^{t}, D^{c} )$ as 
\begin{align}
p(Y^{t}| X^{t}, D^{c} )  \label{eqn:prediction}
&=\int 
p(Y^{t}| X^{t}, \Phi^{\mathcal{B}}  ) 
p(\Phi^{\mathcal{B}} | Z_{\text{Cat}},D^{c} )
\ p(Z_{\text{Cat}}| D^{c}  ) 
\ 
d \Phi^{\mathcal{B}}
d Z_{\text{Cat}}  \nonumber  \\
&\approx \frac{1}{N} \sum_{n=1}^{N} p(Y^{t}| X^{t}, \Phi^{\mathcal{B}}_{(n)}  )  \nonumber
\end{align}
where $Z_{\text{Cat}}$ denotes the categorical latent variable of \cref{eqn:latentcat} for choosing the stationary kernel, $\Phi^{\mathcal{B}}$ denotes the representation of the Bayesian ConvDeepsets $\Phi^{\mathcal{B}}(D^{c})(\bigcdot)$ evaluated on finite grid  $ \{t_m\}_{m=1}^{M}$ (see \cref{eqn:finite_convdeepset}), and $\Phi^{\mathcal{B}}_{(n)}$ denotes the $n$-th random function sampled from $p( \Phi^{\mathcal{B}})$. $p(Y^{t}| X^{t}, \Phi^{\mathcal{B}}_{(n)}  )$ is modeled as 
\begin{align}
p(Y^{t}| X^{t}, \Phi^{\mathcal{B}}_{(n)}  )= N(Y^{t}| \mu_{(n)}(X^{t}), \sigma^{2}_{(n)}(X^{t}) ),
\end{align}
where the $n$-th predictive mean $\mu_{(n)}(X^{t})$ and standard deviation $\sigma^{2}_{(n)}(X^{t})$ on the target inputs $X^{t}$ are obtained by mapping the smoothed features of $X^{t}$ and the $n$-th sample function $\Phi^{\mathcal{B}}_{(n)}$, as explained in \cref{eqn:smoothed}, via the neural network.

\paragraph{Training.} Let $\Theta = \{\theta_{\mathrm{kernels}}, \theta_{\mathrm{p_{\mathrm{nn}}}}, \theta_{\rho},\theta_{\mathrm{pred}}\}$ be the learnable parameters for kernels, the translate invariant network, the CNN, and the network for prediction. Then, using meta-learning, we maximize the following objective w.r.t $\Theta$ :
\begin{align}
 \mathbb{E}_{D^{c},D^{t} \sim p(\mathcal{T})} 
 \bigl[
 \mathcal{L}_{\text{ll}}(\Theta ; D^{c}, D^{t})  - \beta \hspace{.2mm} \mathcal{L}_{\text{reg}}( \Theta ;D^{c}, D^{t}) 
 \bigr]
\label{eqn:tr_objective}
\end{align}
where $\mathcal{L}_{\text{ll}}(\Theta ; D^{c}, D^{t})$ denotes the log likelihood, and $\mathcal{L}_{\text{reg}}(\Theta;D^{c}, D^{t})$ denotes the regularizer that induces the network $\mathrm{p}_{\mathrm{nn}}(D^c)$ in \cref{eqn:latentcat} to assign the reasonable stationary prior depending on context set $D^c$.

The log likelihood $\mathcal{L}_{\text{ll}}(\Theta ; D^{c}, D^{t})$ is expressed as 
\begin{align}
\log{
\left(
 \frac{1}{N} 
 \sum_{n=1}^{N} \exp{ \left(  \sum_{i=1}^{N^{t}} \log{p \left(  y^{t}_{i}| x^{t}_{i}, \Phi^{\mathcal{B}}_{(n)}  \right)}  \right) }   
 \right)  
 },   \nonumber
\end{align}
which can be optimized in end-to-end manner based on the automatic differentiation by \citeb{jang2016categorical,jung2020approximate}.

The regularizer $\mathcal{L}_{\text{reg}}(\Theta;D^{c}, D^{t})$ is expressed as 
\begin{align}
\mathrm{KL} \left( q(Z_{\text{cat}} \ | \  D^{c} )  \ || \  p (Z_{\text{cat}} \ | \  D^{c}, D^{t}  )  \right) \nonumber
\end{align}
where $q(Z_{\text{cat}} \ | \  D^{c} )$  denotes the amortized categorical variable defined in \cref{eqn:latentcat}. $ p (Z_{\text{cat}} \ | \  D^{c}, D^{t} ) $ denotes the posterior distribution of $Z_{\text{cat}}$, which will be used as the prior distribution for training $q(Z_{\text{cat}} \ | \  D^{c} )$. Note that $p (Z_{\text{cat}} \ | \  D^{c}, D^{t} )$ can be computed only during training phase.


The posterior $ p (Z_{\text{cat}} \ | \  D^{c}, D^{t}  )= \mathrm{Cat}(Z_{\text{cat}};\mathrm{p}_{\mathrm{prior}}  ) $ is computed with the parameter $\mathrm{p}_{\mathrm{prior}} \in {\Delta}^{Q-1} $ of which the $q$-th element $(\mathrm{p}_{\mathrm{prior}})_{q} $ is defined as follows: 
\begin{align}
(\mathrm{p}_{\mathrm{prior}})_{q} = 
\frac{\mathrm{exp} \left( \log{p(Y^{t} | X^{t}, D^{c}, k_q)} \ / \  \tau_{0} \right)}
{\sum_{q=1}^{Q} \mathrm{exp} \left( \log{p(Y^{t} | X^{t}, D^{c}, k_q)} \ /\  \tau_{0} \right)}
\end{align}
where $\tau_{0}$ denotes the tempering parameters. The $q$-th marginal likelihood $p(Y^{t} | X^{t},D^{c}, k_q)$, using the $q$-th stationary kernel $k_q$, is computed as
\begin{align}
\hspace{-2.2mm}
p(Y^{t} | X^{t},D^{c}, k_q) = N( Y^{t} ;\hat{\mu}_{q}(X^{t}),\mathrm{Diag}(\hat{\Sigma}_{q}(X^{t})) ) 
\label{eqn:emp-gpposterior}
\end{align}
where $\hat{\mu}_{q}(X^{t}) \in R^{N_t}$ and $\mathrm{Diag}(\hat{\Sigma}_{q}(X^{t})) \in R^{N_t}$ denote the empirical predictive mean and diagonal covariance of the GP posterior distribution on target set $X^{t}$. The sample functions of GP posterior, described in \cref{eqn:randomfunctionalfeature}, are used. The derivation of training loss is explained in Appendix A.3.

%% file: 05-experiments-v05.tex
\vspace{-2mm}
\section{Experiments}
\label{sec:exp}
\vspace{-2mm}

In this section, we validate the following main question: \textb{Does the Bayesian ConvDeepsets alleviate the task ambiguity arising from the small number of context data points ?} We validate this question on the 1-d regression (\cref{subsec:exp1}), multi-channel regression (\cref{subsec:exp2}), and large-sized image completion (\cref{subsec:exp3}). The detailed setting and additional results are given in Appendix B-E.

\input{05-experiments-v05-part01}

\input{05-experiments-v05-part02}

\input{05-experiments-v05-part03}

%% file: 05-experiments-v05-part01.tex



\subsection{Single-Channel Regression}
\label{subsec:exp1}

\begin{figure*}[h]
\subfloat[\label{fig:1d-singletask-a}  Spectral density]
{\includegraphics[width=0.19\linewidth  ,height=2.5cm ]{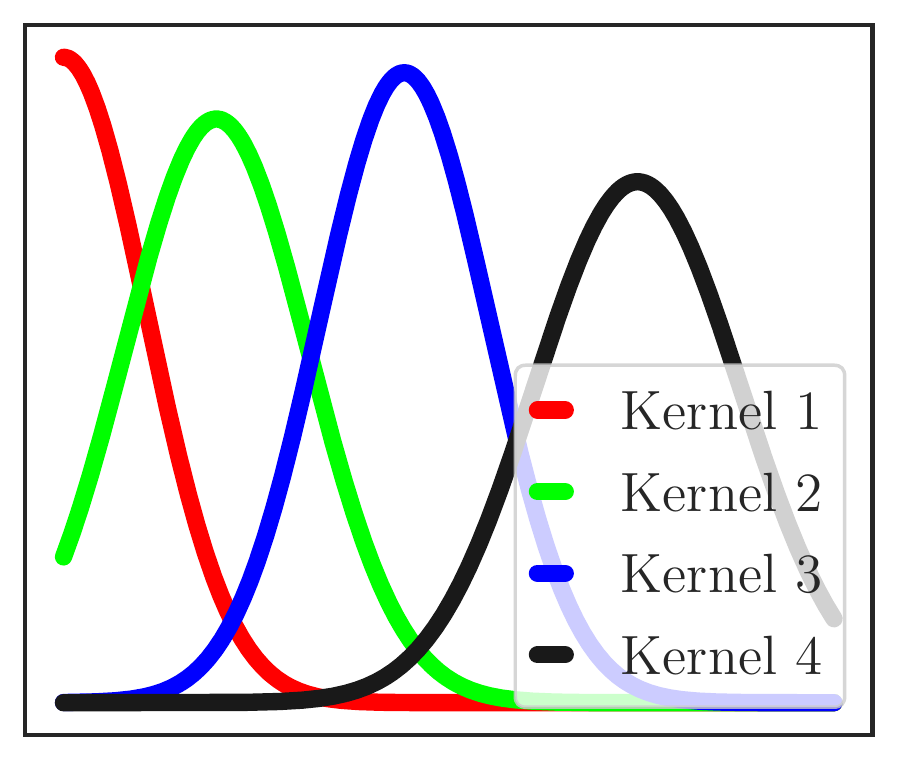}} \hspace{0.1mm}
\subfloat[\label{fig:1d-singletask-b} RBF]
{\includegraphics[width=0.19\linewidth ,height=2.5cm ]{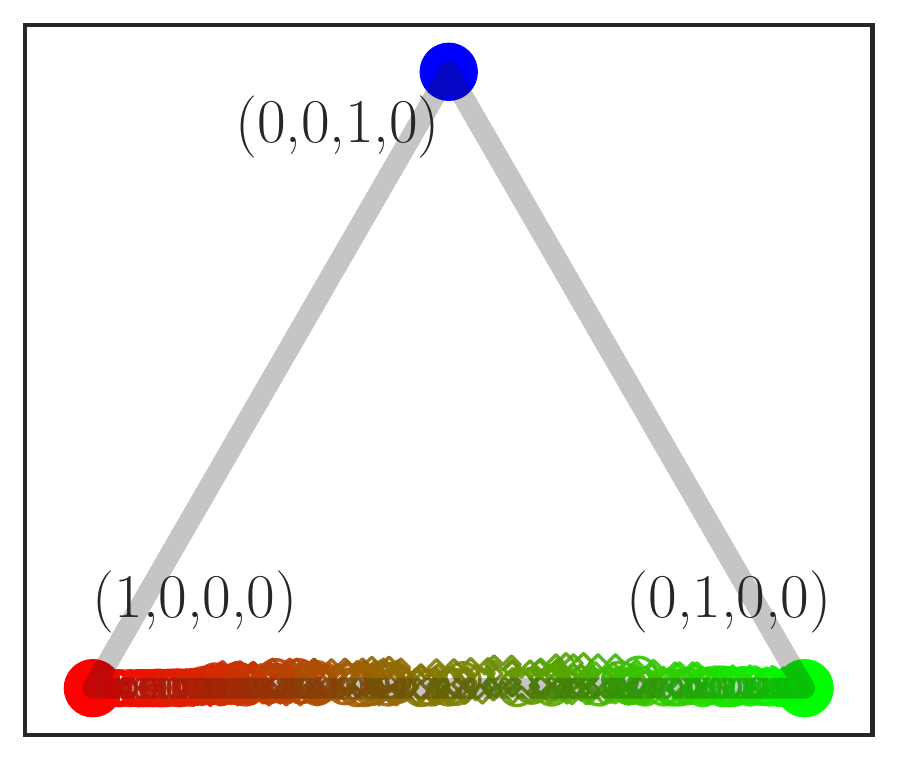}} \hspace{0.1mm}
\subfloat[\label{fig:1d-singletask-c} Matern-${\frac{5}{2}}$]
{\includegraphics[width=0.19\linewidth ,height=2.5cm ]{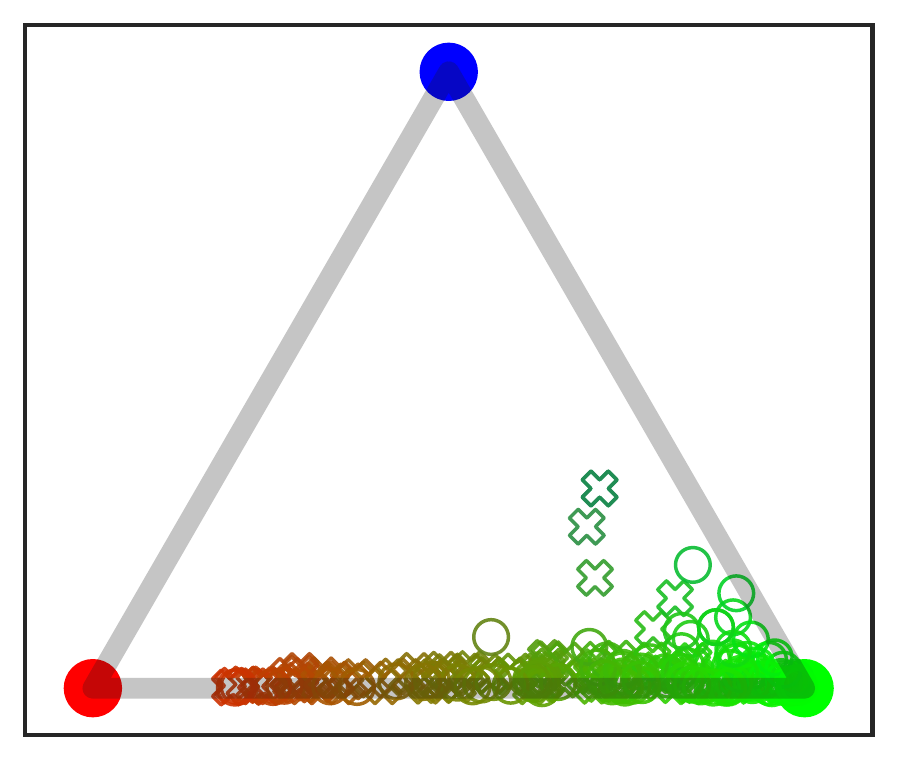}} \hspace{0.1mm}
\subfloat[\label{fig:1d-singletask-d} Weakly Periodic]
{\includegraphics[width=0.19\linewidth ,height=2.5cm ]{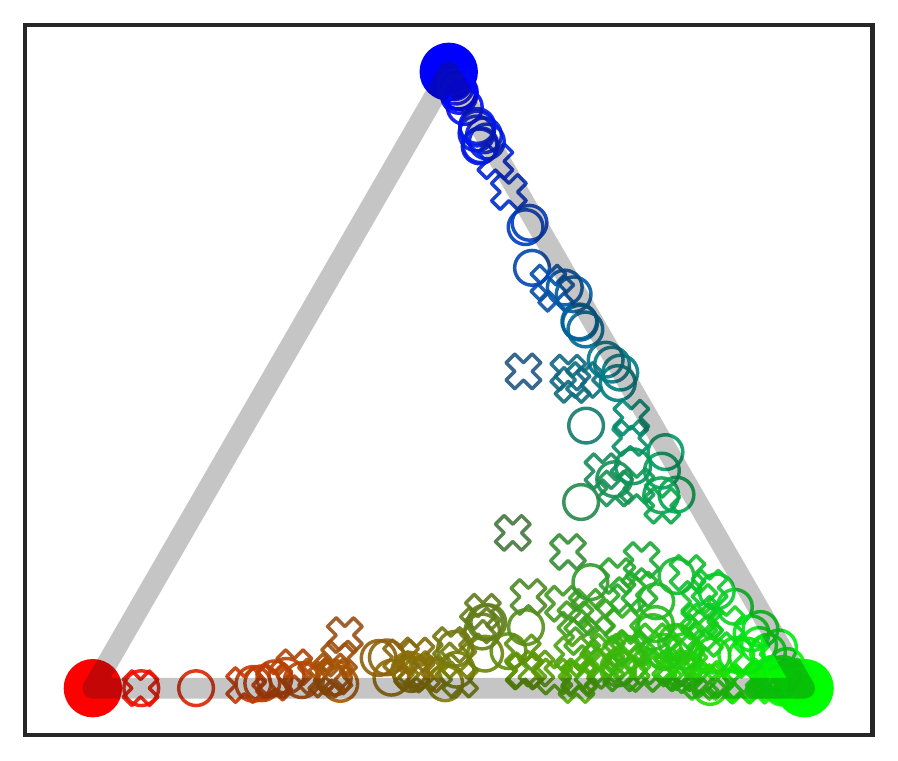}} \hspace{0.1mm}
\subfloat[\label{fig:1d-singletask-e} Sawtooth]
{\includegraphics[width=0.19\linewidth ,height=2.5cm ]{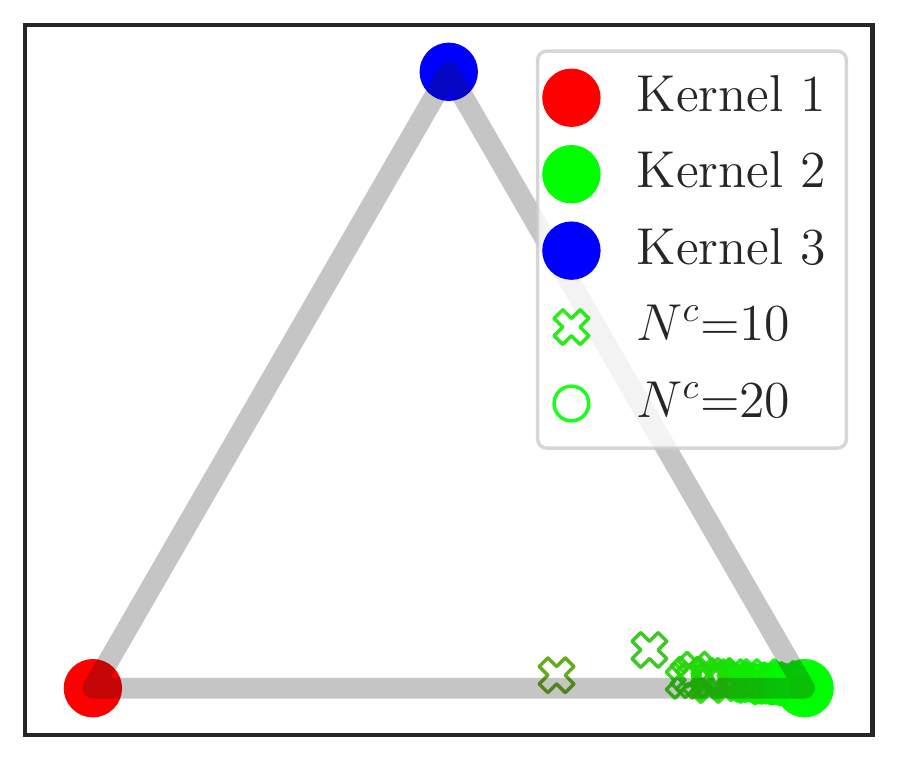}}  
\vspace{1mm}
\subfloat[\label{fig:1d-singletask-f} Predictive distribution of each baseline on each process ($N^c=15$) ]
{\includegraphics[width=0.99\linewidth  ]{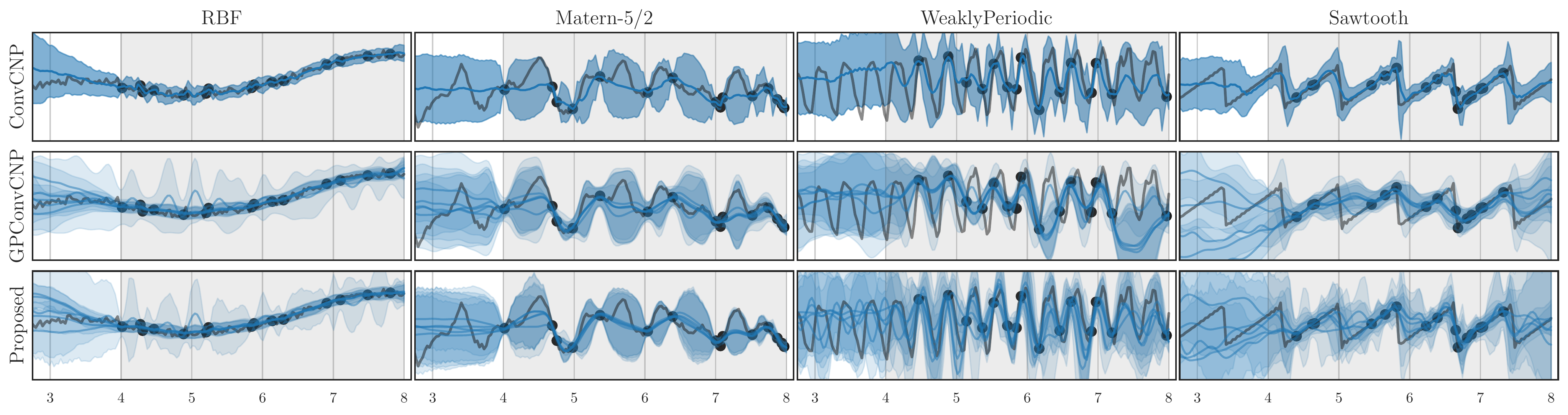}}
\caption{Stochastic processes modeling beyond training range (gray region $[4,8]$): \cref{fig:1d-singletask-a} denotes the trained spectral density of \cref{eqn:basiskernel} with $Q=4$ kernels. \cref{fig:1d-singletask-b,fig:1d-singletask-c,fig:1d-singletask-d,fig:1d-singletask-e} denote the output of $\mathrm{p}_{\mathrm{nn}}(D^{c})$ for 128 context set $\{D^{c}_{n}\}_{n=1}^{128}$ of each process. \cref{fig:1d-singletask-f} shows the predictive distributions of the baselines on each process. 
}

\label{fig:1d-singletask-prediction}
\end{figure*}
\vspace{-2mm}

We conduct the 1-d regression task based on meta learning framework; given each context set $D^{c}=\{(x^{c}_{n},y^{c}_{n})\}_{n=1}^{N^c}$, the model constructs the predictive distribution on the target set $D^{c}=\{(x^{t}_{n},y^{t}_{n})\}_{n=1}^{N_t}$. Specially, we consider each task with a small number of context points $N^{c}$ during training to investigate how $N^{c}$ data points affects the representation of ConvDeepsets, and the performance of the NP models. For each task, we set the number of context points $N^{c} \sim U(5,25)$ and target points $N^{t} \sim  U(N^{c} ,50)$, using half the number of context data points compared to the setting of the ConvCNP \citep{gordon2019convolutional}. In addition, we consider different task diversities to investigate whether the task-dependent prior can be effective even when $N^c$ is insufficiently small, making it difficult to recognize each task.

For training and validation, we use the task generated on $[0,4]$ (training range), and then evaluate the trained models with the tasks on $[4,8]$ to check the TE in \cref{eqn:def1,eqn:te_probsense}.
\vspace{-3mm}
\paragraph{Dataset.} We use 4 different stochastic processes used in \citeb{gordon2019convolutional}: GP using the RBF, Matern-${\frac{5}{2}}$, and Weakly periodic kernel, and sawtooth process that is not GP. For the task diversity, every task is generated by one of processes randomly, and then used for training; this setting is different to that conducted in \citeb{gordon2019convolutional,foong2020meta}, where the NP models are trained and evaluated on each stochastic process independently.
\vspace{-3mm}
\paragraph{Baselines.} We use the Attentive NP (ANP) \citeb{kim2018attentive}, ConvCNP \citeb{gordon2019convolutional}, ConvLNP \citeb{foong2020meta},  GPConvCNP \citeb{petersen2021gp} that uses the sample function of GP posterior with RBF kernel.


\paragraph{Results.}

\vspace{-2mm}

\begin{figure}[H]
\centering
\subfloat[RBF]
{\includegraphics[width=0.24\linewidth]{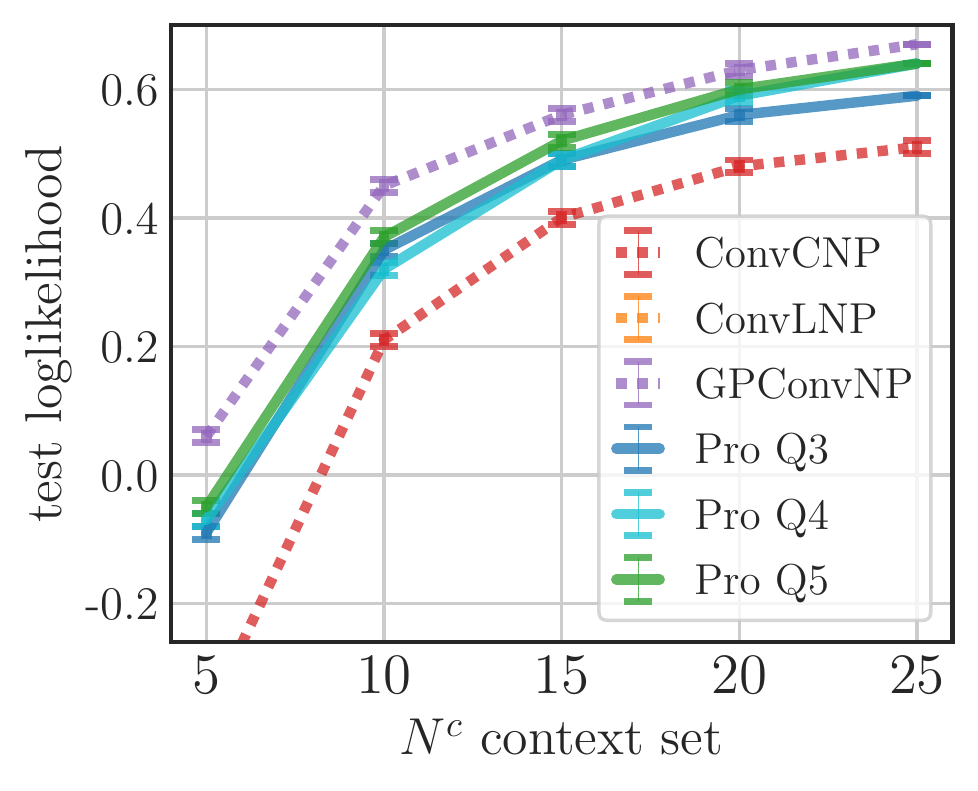}}   \hspace{0.1mm} 
\subfloat[ Matern-${\frac{5}{2}}$] 
{\includegraphics[width=0.24\linewidth]{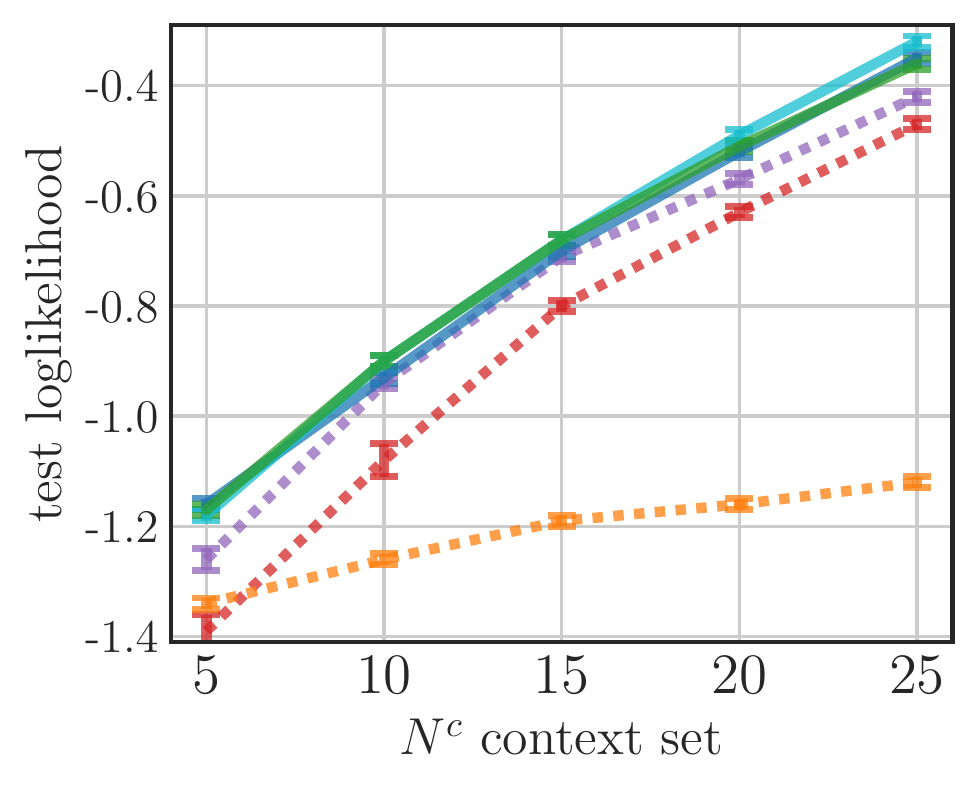}}   
\hspace{2mm} 
\subfloat[Weakly periodic]
{\includegraphics[width=0.24\linewidth]{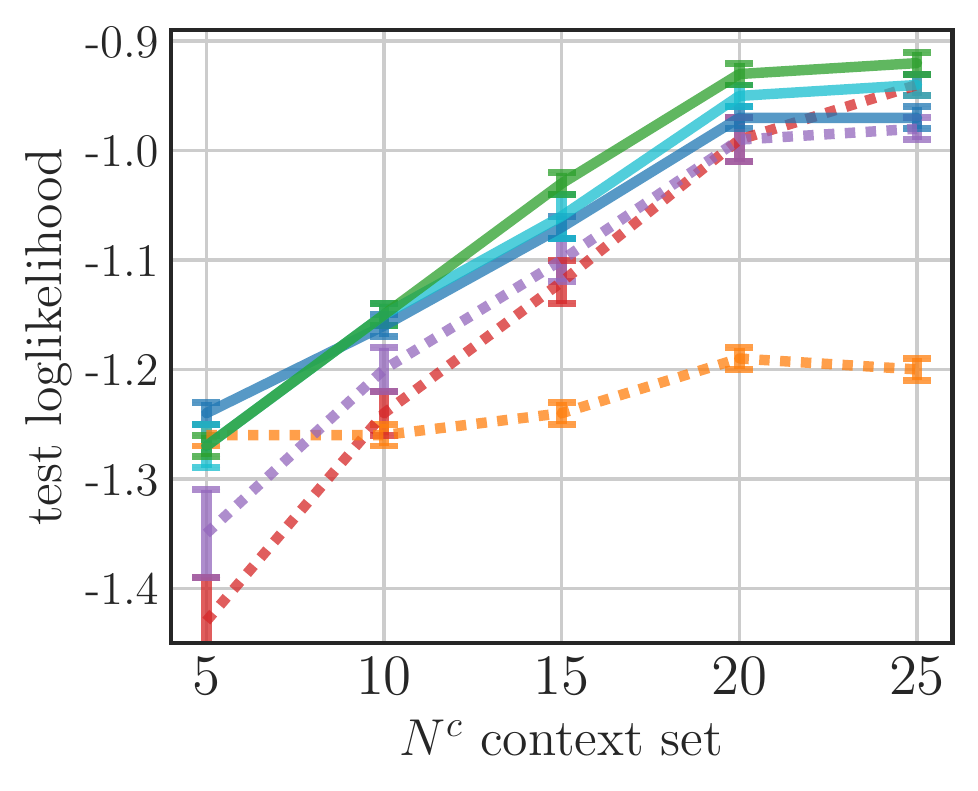}}  \hspace{0.1mm} 
\subfloat[Sawtooth]
{\includegraphics[width=0.24\linewidth]{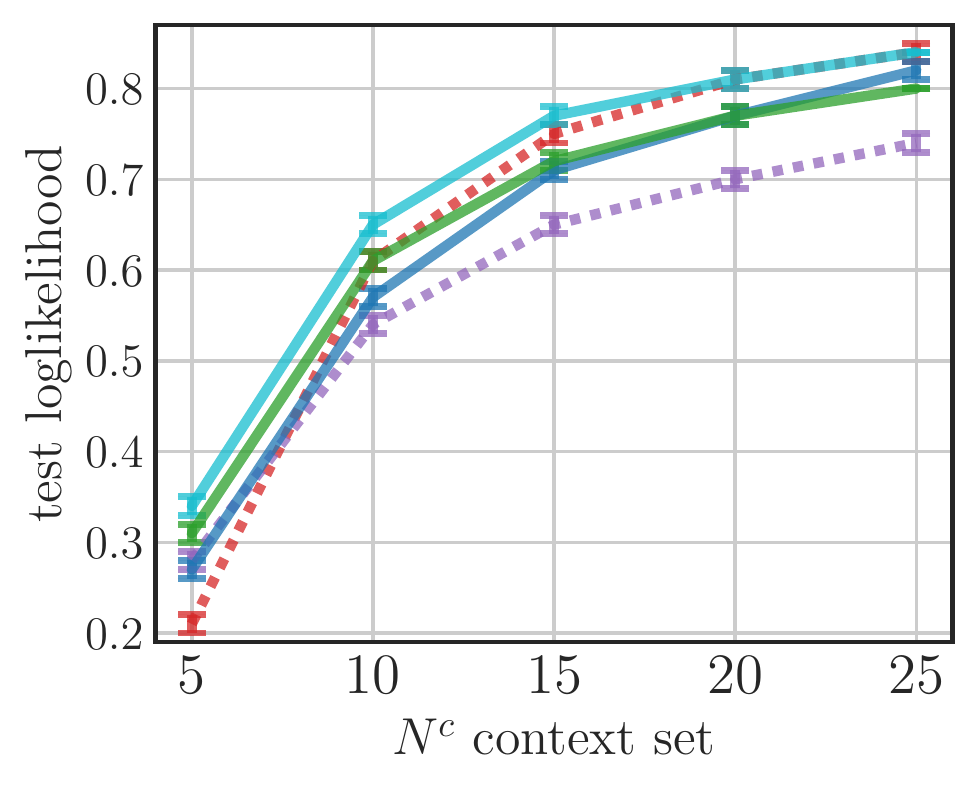}}   
\caption{Prediction result of each processes on varying $N^c$.   }
\label{fig:exp5-1:smallset_comparison}
\end{figure}
\vspace{-1mm}

\cref{fig:exp5-1:smallset_comparison} describes the mean and one-standard error of the log likelihood for $1024$ tasks (beyond training range); for evaluation, we set varying number of context data points $N^c \in \{5,10,15,20,25\}$ and target points $N^{t}=50$ to investigate the effectiveness of the proposed method on varying  $N^c$. We obtain the following observation:
\begin{itemize}[leftmargin=1em]
    \vspace{-1.1mm}
    \item The NP model, that uses Bayesian ConvDeepsets ($Q\in\{3,4,5\}$) predicts the target set most accurately (high log likelihood) for the tasks of each process; for small context points $N^{c}\in \{5,10\}$, the proposed method outperforms ConvCNP with a large margin. This margin decreases as the number of context points $N^{c}$ increases.
    
    \vspace{-1.1mm}
    \item GPConvCNP and ConvCNP show good prediction performance on specific process (GPConvCNP: RBF, ConvCNP: Sawtooth), whereas the proposed method show good prediction performance on all processes. ANP shows the bad generalization on tasks beyond the training range, and thus can not be reported together.
\end{itemize}
\vspace{-1mm}
Additionally, we conduct similar experiments with a larger number of context data points $N^{c} \sim U(10,50)$ in training, and confirm that the gap between the propose model and ConvCNP is reduced in Appendix B.4.
\vspace{-1.5mm}

\cref{fig:1d-singletask-a} describes the spectral densities of \cref{eqn:basiskernel} for the trained kernels ($Q=4$).
\cref{fig:1d-singletask-b,fig:1d-singletask-c,fig:1d-singletask-d,fig:1d-singletask-e} show the output of $\mathrm{p}_{\mathrm{nn}}(D^{c})$, i.e. the parameters of $ \mathrm{Cat} \left( Z_{\text{cat}} \ ; \ \mathrm{p}_{\mathrm{nn}}(D^{c}) \right)$ in \cref{eqn:latentcat}, for 128 context sets $\{D^{c}_{n}\}_{n=1}^{128}$ of each process. We see that the trained model imposes the stationary prior differently depending on the tasks of each process. \cref{fig:1d-singletask-f} shows the prediction results for one task ($N_c=15$); each column and row shows the task of each process and the NP models. We see that the prediction on RBF and WeaklyPeriodic task is affected by different stationary prior of kernel 1 and 3.

%% file: 05-experiments-v05-part02.tex
\vspace{-1mm}
\subsection{Multi-Channel Regression}

\label{subsec:exp2}

\vspace{-2mm}
\paragraph{Experiment setting.} We conduct the multi-channel regression task. We consider different task diversities to investigate the effectiveness of the task-dependent prior especially for a small number of  context points $N_c \sim \mathcal{U}([5,25])$.

\vspace{-1mm}

\paragraph{Dataset.}
We use the 3-channels sinusoidal process of which the $i$-channel function $f_{i}$ is represented as 
\vspace{-1mm}
\begin{align}
 f_{i}(t) = A_i\sin{ \left( 2\pi (w_i + \theta_i) (t - \phi_i) \right)} + \epsilon,   
 \hspace{2mm} \epsilon \sim N(0,0.1^2), \nonumber
\end{align}
where $\{A_i,w_i,\phi_i\}_{i=1}^{3}$ denotes the $i$-channel amplitude, frequency, and phase parameters. For the task diversity, one process is generated from the fixed amplitude $A_i$, frequency $w_i$, and a varying phase parameter $\phi_i$ that is sampled differently for each task. Another process is generated from the varying amplitude, frequency, and phase, which is regarded to have the high diversity. For the details, see Appendix C.2.

\begin{figure*}[h]
\centering
\subfloat[\label{fig:nd-multitask-a} Sinusoidal-phase ]
{\includegraphics[width=0.23\linewidth]{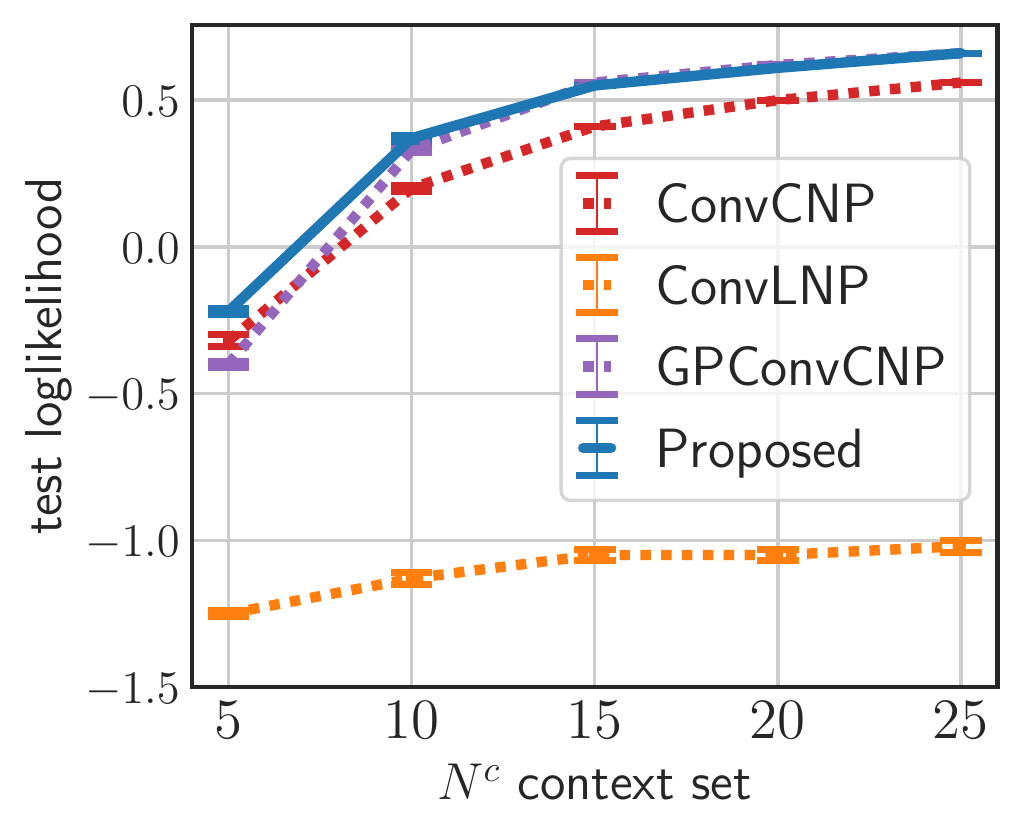}} 
\hspace{.1mm}
\subfloat[\label{fig:nd-multitask-b} Sinusoidal-all ]
{\includegraphics[width=0.23\linewidth]{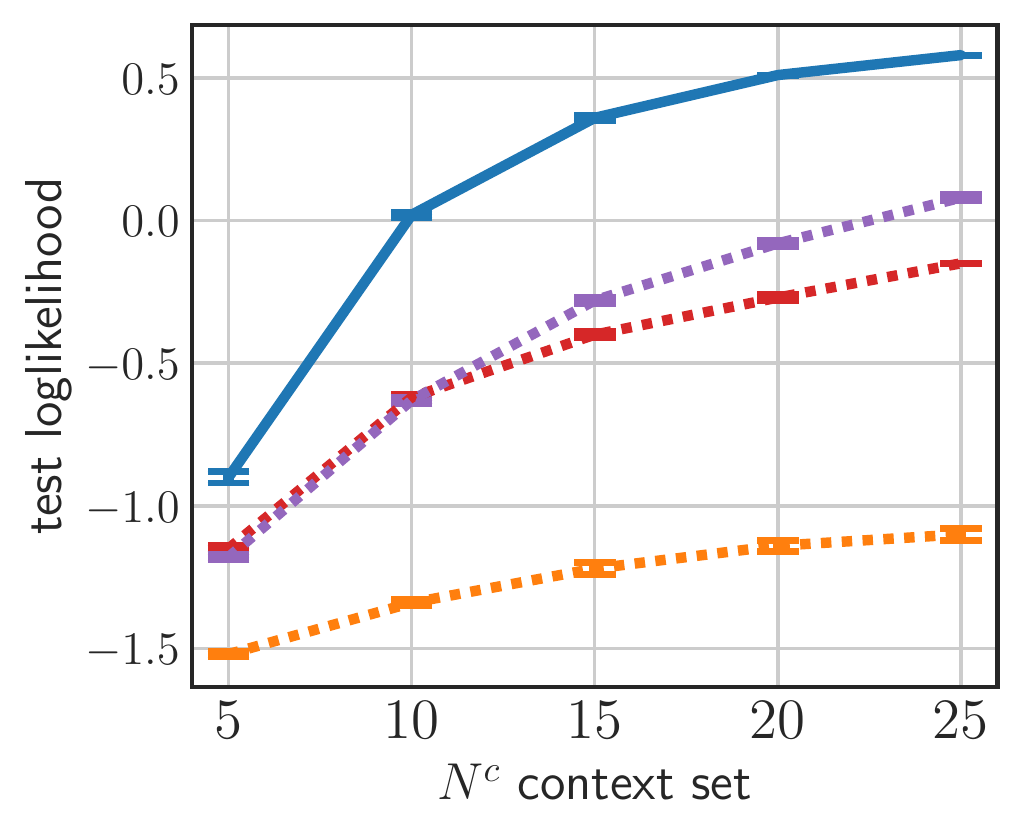}} 
\hspace{3mm}
\subfloat[\label{fig:nd-multitask-c} Spectral density]
{\includegraphics[width=0.23\linewidth]{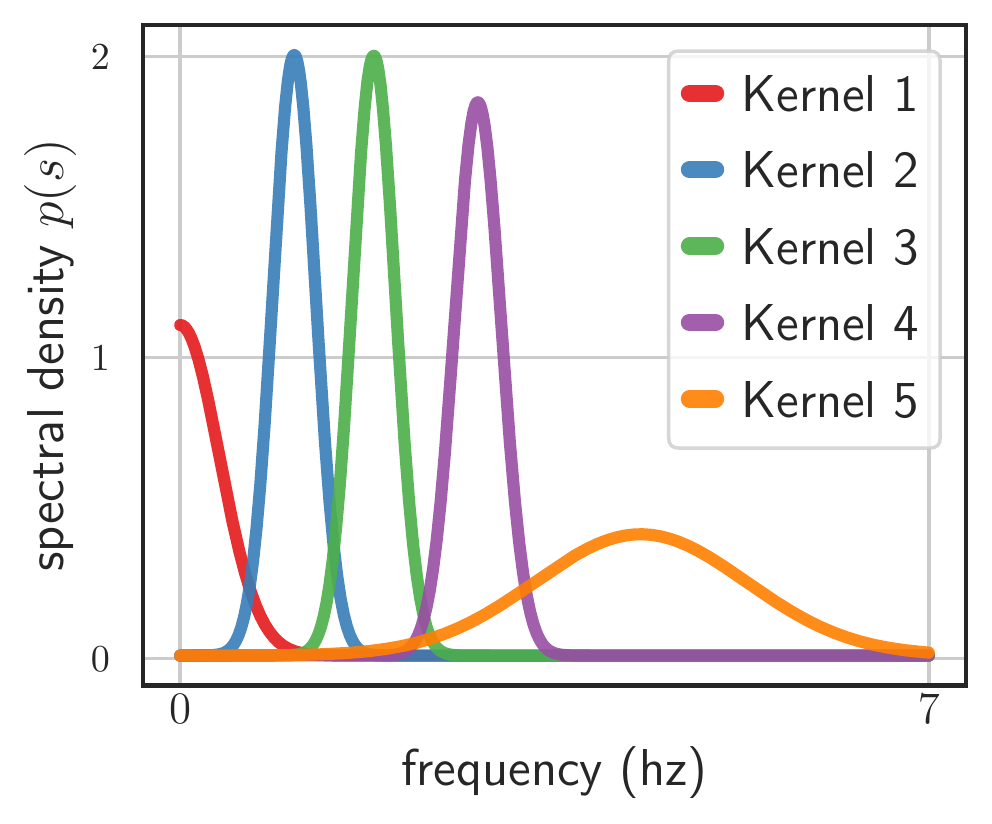}}
\hspace{.1mm}
\subfloat[\label{fig:nd-multitask-d} Task-dependent prior]
{\includegraphics[width=0.23\linewidth]{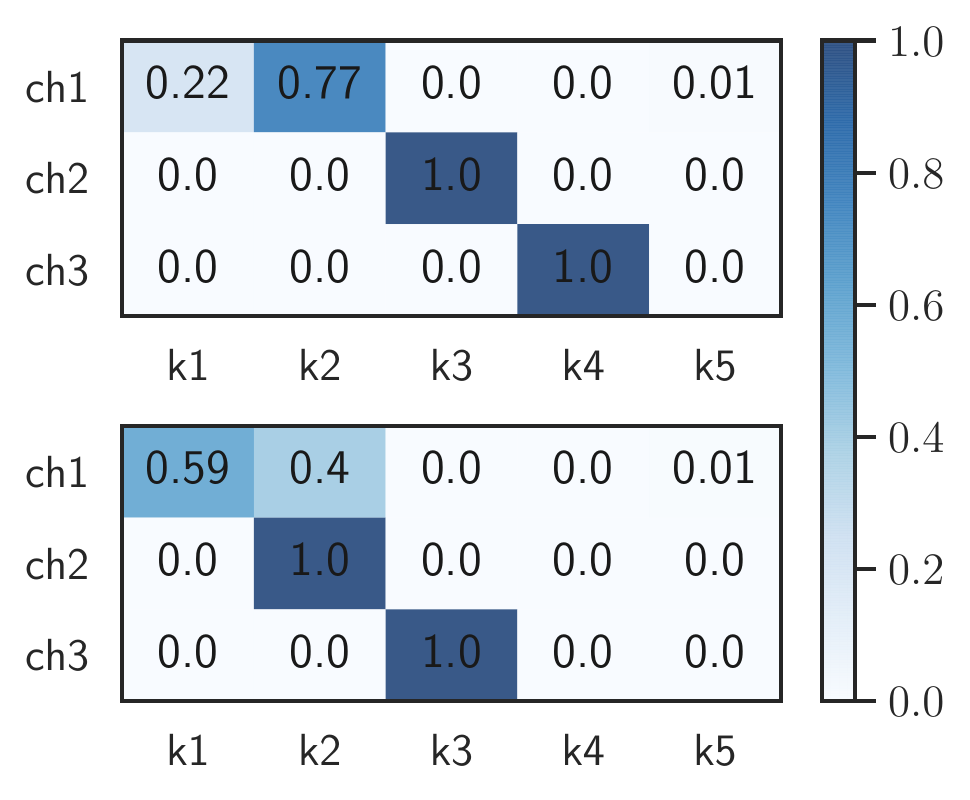}}
\vspace{2mm}
\subfloat[\label{fig:nd-multitask-e} Prediction for 2 different tasks ($N^{c}=10$) of the sinusoidal-varying  ]
{\includegraphics[width=0.98\linewidth]{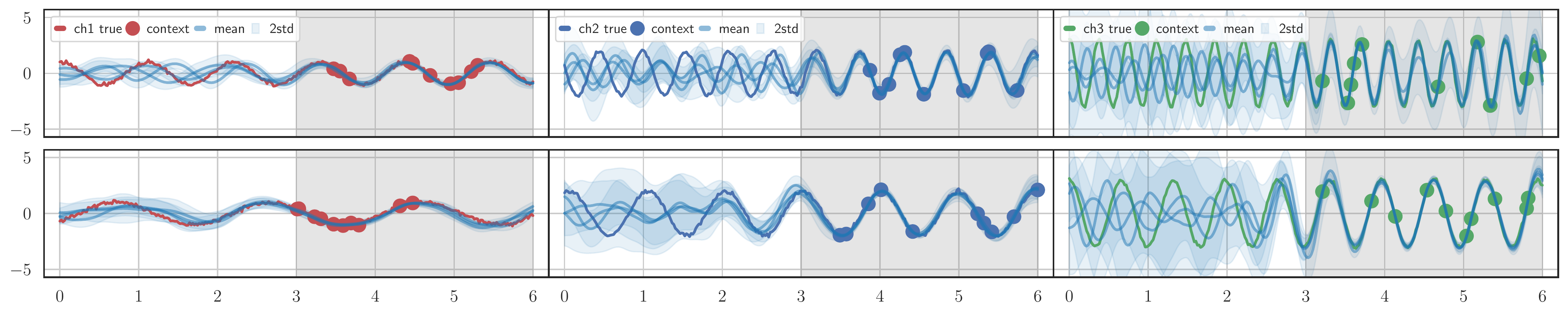}} 
\caption{3-channel sinusoidal processes modeling: \cref{fig:nd-multitask-a,fig:nd-multitask-b} denotes test likelihood on sinusoidal processes having different task diversity. \cref{fig:nd-multitask-c} denote the trained spectral density of \cref{eqn:basiskernel} for the Sinusoidal-all process, and \cref{fig:nd-multitask-d,fig:nd-multitask-e} show the chosen prior and corresponding prediction for 2 tasks; as the context sets having small frequency characteristics are given (from first to second row in \cref{fig:nd-multitask-e}), the stationary prior is imposed differently (from top to bottom in \cref{fig:nd-multitask-d})
}
\label{fig:nd-multitask-pred}
\end{figure*}


\vspace{-2mm}
\paragraph{Results.}

\cref{fig:nd-multitask-a,fig:nd-multitask-b} describes the mean and one-standard error of the log likelihood for 1024 tasks (beyond training range); \cref{fig:nd-multitask-a} shows the result of the less diversity task (varying phase), and \cref{fig:nd-multitask-b} corresponds to that of the high diversity (varying amplitude, frequency, and phase). These figures show that the proposed method could model processes well on both less and more diverse tasks. 

\cref{fig:nd-multitask-c} shows the trained spectral density of $\{k_{q}\}_{q=1}^{5}$ in \cref{eqn:basiskernel} on high diversity task. \cref{fig:nd-multitask-d} denote the parameters $\mathrm{p}_{\mathrm{nn}}(D^{c})$ for two tasks (top and bottom), and \cref{fig:nd-multitask-e} shows their corresponding predictions. This shows how the task-dependent prior is imposed, and affect the prediction on target set. We confirm similar results on other datasets (GP using the MOSM kernel \citeb{parra2017spectral}) in Appendix C.4

%% file: 05-experiments-v05-part03.tex

\vspace{-1mm}
\subsection{Image Completion} 
\label{subsec:exp3}

\begin{figure*}[h]
\centering
{\includegraphics[width=0.99\linewidth,height=5.0cm]{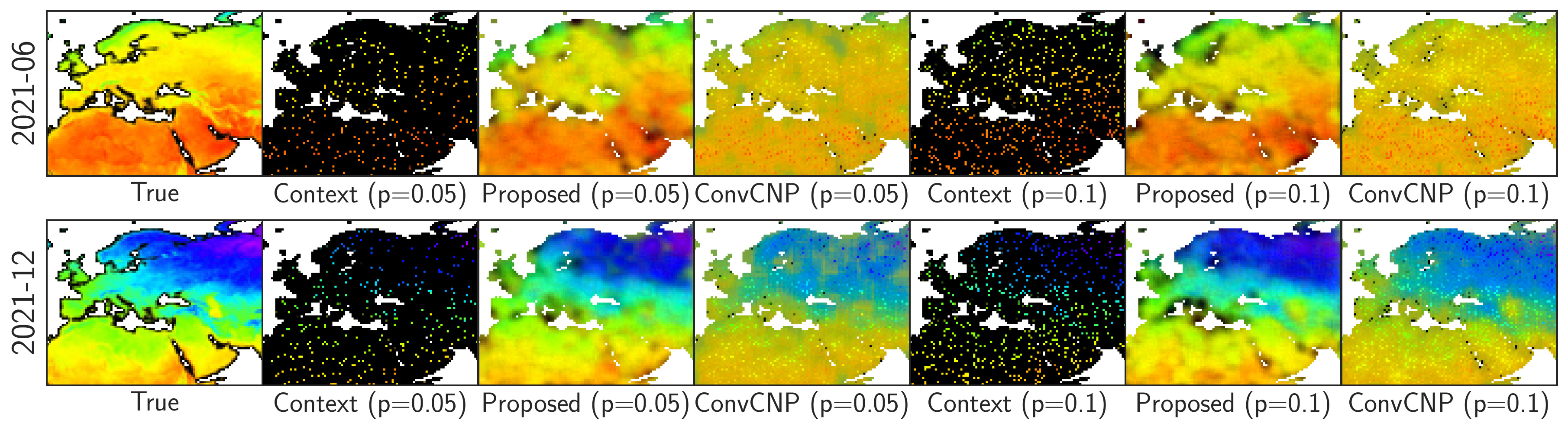}}  
\caption{Prediction results for two Europe temperatures (June 2021 and December 2021) that is out of training sets.}
\label{fig:image_completion-usa}
\end{figure*}

\vspace{-1mm}
\paragraph{Experiment setting.} 

We conduct the image completion task with a monthly land surface temperature set used in \citeb{remes2017non} because the temperature has the different local stationarity depending on the month. Each task is to predict the image of the temperature when only partial pixel values are given. The temperature of North America ($53{\times}115{\times}3$) and Europe ($78{\times}102{\times}3$) are used for the training (2018 - 2020), and for the test (2021). For each task, we set that $N^c$ context points and $N^{t}$ target points is proportional to the size of image by choosing a small context rate $p \in \{.01,.05,.10,.20\}$ randomly with probability $[4/10,3/10,2/10,1/10]$. This is indented to investigate how NP models predict the image of temperature when NP models are trained with each context set $D^{c}$ having the the small number $N^{c}$ of context points.

\vspace{-2.5mm}
\paragraph{Results.} 
\vspace{-2mm}

\begin{figure}[H]
\subfloat[\label{fig:imag-a} North America (in-train)]
{\includegraphics[width=0.24\linewidth,height=3cm]{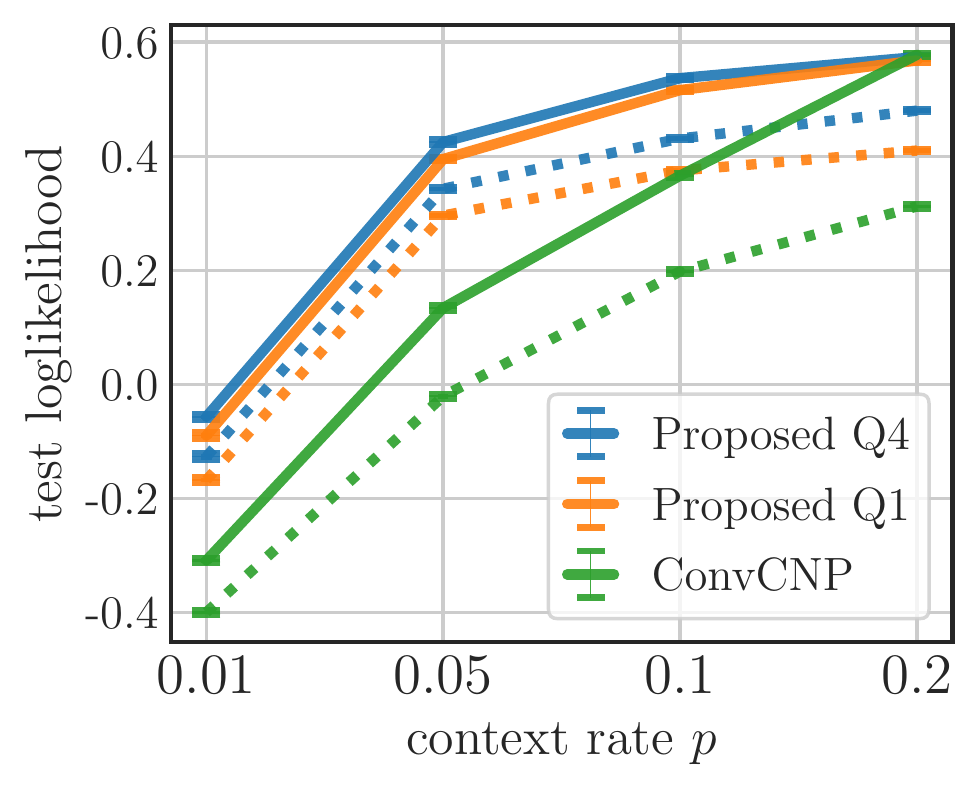}} \hspace{.5mm}  
\subfloat[\label{fig:imag-b} North America (out-train)]
{\includegraphics[width=0.24\linewidth,height=3cm]{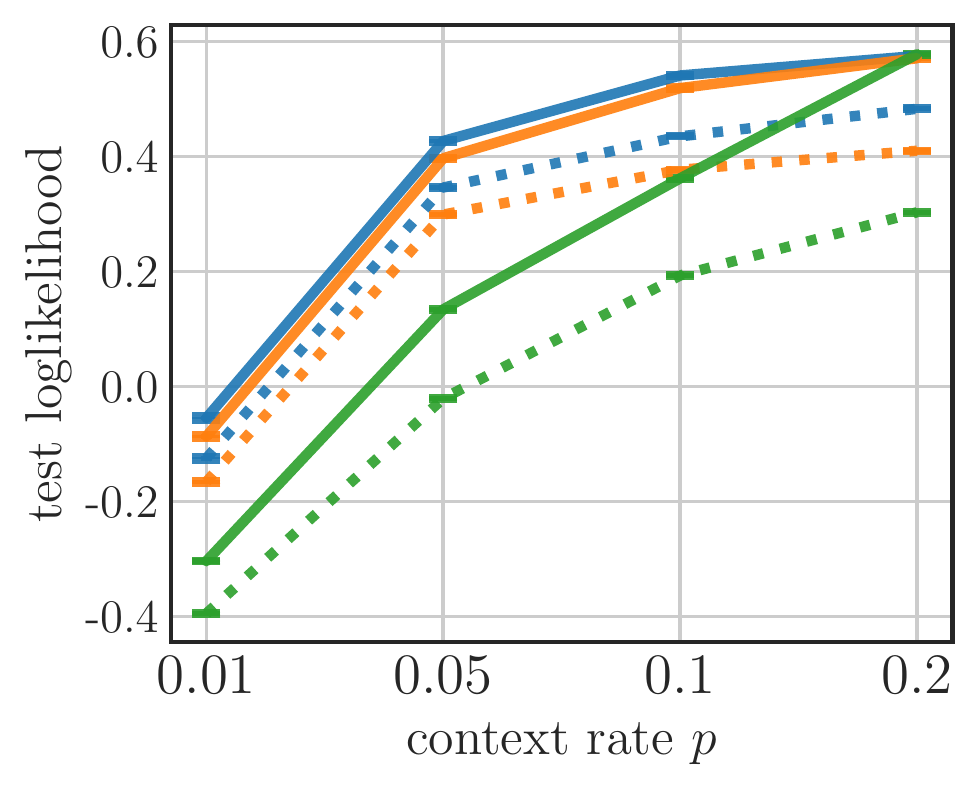}} 
\hspace{2mm}
\subfloat[\label{fig:imag-c} Europe (in-train)]
{\includegraphics[width=0.24\linewidth,height=3cm]{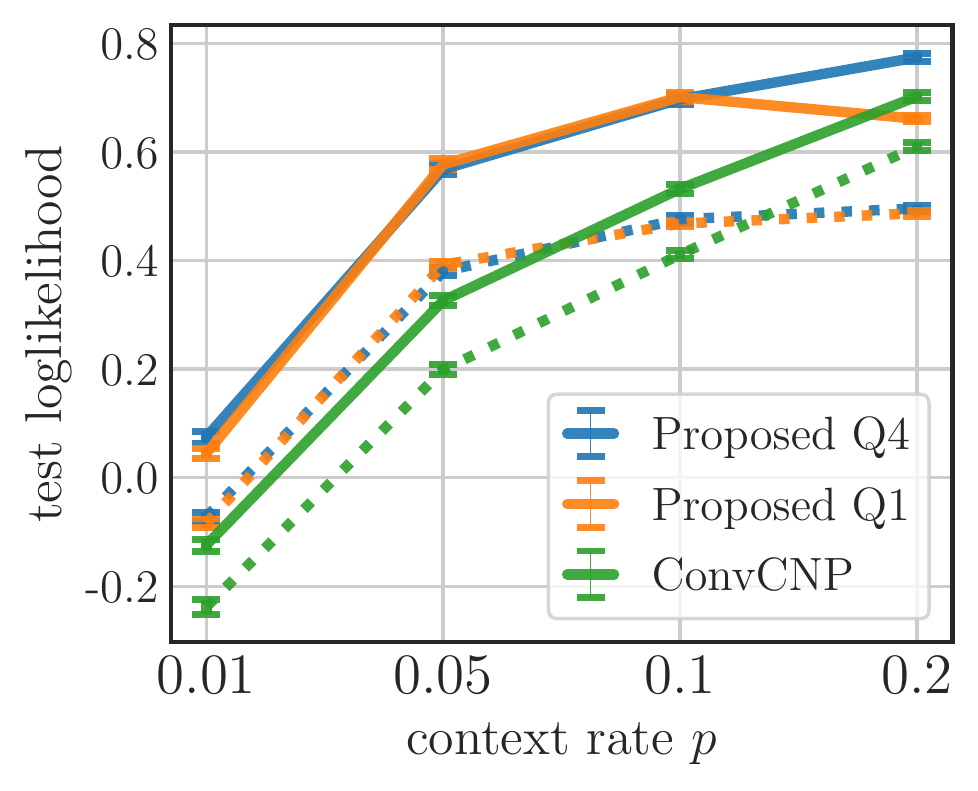}} \hspace{.5mm}  
\subfloat[\label{fig:imag-d} Europe (out-train)]
{\includegraphics[width=0.24\linewidth,height=3cm]{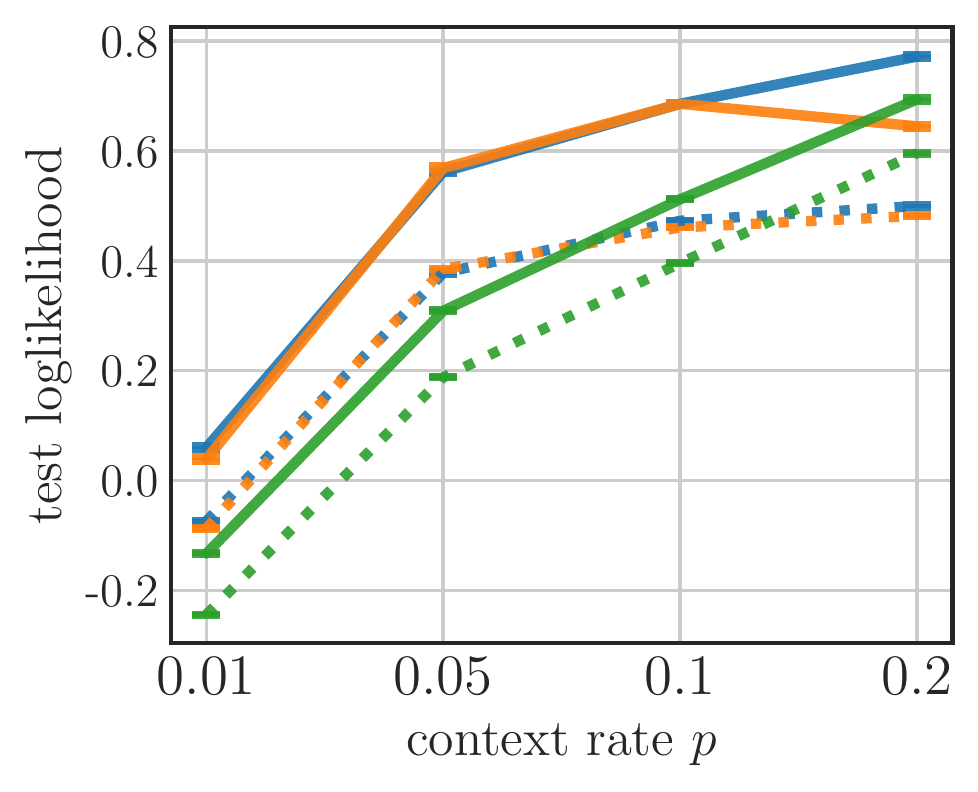}} 
\caption{{Prediction Results of each 256 tasks over context rate $p \in \{.01,.05,.10,.20\}$ and fixed target rate $p=.50$.}}
\vspace{-3mm}
\end{figure}

\cref{fig:imag-a,fig:imag-c} describes the log likelihood for the North America and Europe datasets (2018 - 2020), respectively; the dotted ($\cdot$) and straight (-) line denote the results obtained at 5000 training tasks and 10000 training tasks, respectively.
\cref{fig:imag-b,fig:imag-d} corresponds to the result of out-of-training set (2021). We see that the proposed approximation ($Q=4$) outperforms ConvCNP on small $p\in \{.01,.05\}$ and 5000 training tasks. As the number of tasks increases for training, ConvCNP reduces the performance gap with proposed method. We guess that the accumulated tasks during training cover the task diversity of the given datasets, and enable ConvCNP to model most of target tasks.  \cref{fig:image_completion-usa} shows the targeted image of the Europe (June and December of 2021), context set with $p\in\{.05,0.10\}$, and the predictive mean of each model, which are obtained by 10000 training tasks.

\vspace{-3mm}

%% file: 04-relatedwork-v05.tex
\section{Discussion}


%

\paragraph{Comparison with ConvLNP.} Convolutional latent neural process (ConvLNP) \citeb{foong2020meta} considers the latent variable into ConvDeepsets for relaxing the conditional independence assumption of the ConvCNP. However, ConvLNP could have the same issue with ConvCNP because it uses the ConvDeepsets. Also, training the ConvLNP is known to be difficult in Section 7, \citeb{gordon2021advances}; we believe this difficulty arises because the latent variables are constructed from the ambiguous representation of ConvDeepsets, and the randomness of the latent variable prevents the NP model from fitting the target as well. On the other hand, the proposed method uses the multiple data representations of \cref{eqn:randomfunctionalfeature} via the task-dependent prior, and use them for modeling the task.


\vspace{-2mm}
\paragraph{Comparison with GPConvCNP.}

GPConvCNP \citeb{petersen2021gp} uses the sample function of GP Posterior (RBF kernel) as the functional representation. This is similar to our work. However, our work focuses more on generalizing the ConvDeepsets in Bayesian way; 1) we propose how to impose the task-dependent prior, 2) prove that the Bayesian ConvDeepsets still holds the TE as shown in \cref{pro:Bayes_convdeepset}, and 3) present the scalable approximation for large-sized set.

\vspace{-2mm}
\paragraph{Limitation.} The Bayesian ConvDeepsets employs the task-dependent stationary prior to remedy the issue of ConvDeepsets. However, if a targeted process could not be modeled by stationary process, the proposed method may not be helpful to model targeted process. Also, if many context points and tasks are allowed for training at the cost of data collection, the advantage of using stationary prior could be insignificant, and so ConvCNP is recommended.



